\documentclass[twoside,11pt]{article}

\usepackage{jmlr2e}
\usepackage{float}
\usepackage{multirow}
\usepackage{amsmath}
\usepackage{caption}
\usepackage{subcaption}
\usepackage{rotating}
\usepackage{epstopdf}
\usepackage{color}
\floatstyle{ruled}
\newfloat{algorithm}{tbp}{loa}
\providecommand{\algorithmname}{Algorithm}
\floatname{algorithm}{\protect\algorithmname}

\ShortHeadings{Marginal Pseudo-Likelihood Learning of Markov Network Structures}{Pensar, Nyman, Niiranen and Corander}
\firstpageno{1}

\begin{document}

\title{Marginal Pseudo-Likelihood Learning of \\ Markov Network Structures}

\author{\name Johan Pensar\vspace{0.1cm} \email johan.pensar@abo.fi \\
	   \name Henrik Nyman \email henrik.nyman@abo.fi \\
       		\addr Department of Mathematics and statistics\\
      		 {\AA}bo Akademi University\\
      		 20500 Turku, Finland
		       \AND
	   \name Juha Niiranen\vspace{0.1cm} \email juha.niiranen@helsinki.fi \\	
	   \name Jukka Corander \email jukka.corander@helsinki.fi \\
      		 \addr Department of Mathematics and statistics\\
       		University of Helsinki\\
       		00014 Helsinki, Finland
}

\editor{}

\maketitle

\begin{abstract}
Undirected graphical models known as Markov networks are popular for a wide variety of applications ranging from statistical physics to computational biology. Traditionally, learning of the network structure has been done under the assumption of chordality which ensures that efficient scoring methods can be used. In general, non-chordal graphs have intractable normalizing constants which renders the calculation of Bayesian and other scores difficult beyond very small-scale systems. Recently, there has been a surge of interest towards the use of regularized pseudo-likelihood methods for structural learning of large-scale Markov network models, as such an approach avoids the assumption of chordality. The currently available methods typically necessitate the use of a tuning parameter to adapt the level of regularization for a particular dataset, which can be optimized for example by cross-validation. Here we introduce a Bayesian version of pseudo-likelihood scoring of Markov networks, which enables an automatic regularization through marginalization over the nuisance parameters in the model. We prove consistency of the resulting MPL estimator for the network structure via comparison with the pseudo information criterion. Identification of the MPL-optimal network on a prescanned graph space is considered with both greedy hill climbing and exact pseudo-Boolean optimization algorithms. We find that for reasonable sample sizes the hill climbing approach most often identifies networks that are at a negligible distance from the restricted global optimum. Using synthetic and existing benchmark networks, the marginal pseudo-likelihood method is shown to generally perform favorably against recent popular inference methods for Markov networks.
  
\end{abstract}


\section{Introduction}
Markov networks represent a ubiquitous modeling framework for multivariate systems, with applications ranging from statistical physics to computational biology and sociology \citep[see][]{Lauritzen1996,Koller2009}. However, statistical inference for such models is in general challenging, both regarding estimation of parameters and learning structure of the network. Under the assumption of chordality it is possible to use a closed-form factorization of a distribution with respect to a Markov network, however, in non-chordal cases the normalizing factor (or partition function) of these distributions  is intractable beyond toy-sized systems. Since the chordality assumption is restrictive and may seriously bias learning of the dependencies among variables, considerable interest has been targeted towards making also non-chordal networks tractable for applications. A revival of interest has in particular arisen from the need to consider high-dimensional models in a 'large \textit{p}, small \textit{n}' setting \citep{Lee2006,Hofling2009,Ravikumar2010,Aurell2012,Aurell2013}.

In physics, Markov network models have traditionally been fitted using the mean-field approximation, which has only recently started to become superceded by more elaborate approaches, such as the pseudo-likelihood method \citep{Aurell2012,Aurell2013}. The pseudo-likelihood approach was originally motivated by the difficulties of maximizing the likelihood function for lattice models \citep{Besag1972} and it simplifies the model fitting by a factorization of the likelihood over local neighborhoods of the random variables involved in the modeled system.  

High-dimensional Markov networks usually necessitate some form of regularization to make the pseudo-likelihood estimation problem feasible to solve. Some of the currently available methods necessitate the use of a tuning parameter to adapt the level of regularization for a particular dataset. The value of the tuning parameter can then be optimized for example by cross-validation. Here we introduce a Bayesian version of the pseudo-likelihood approach to learn the structure of a Markov network without assuming chordality.  Our method enables an automatic regularization of the resulting model complexity through marginalization over the nuisance parameters in the model. 

The structure of the remaining article is as follows. In the next section the basic properties of Markov networks are reviewed and the structure learning problem is formulated in Section 3. In Section 4, we introduce the marginal pseudo-likelihood (MPL) score and prove consistency of the corresponding structure estimator. Algorithms for optimizing the MPL score for a given dataset are derived in Section 5 and the penultimate section demonstrates the favorable performance of our method against other popular recent alternatives. The last section provides some additional remarks and conclusions.  

\section{Markov networks}\label{sec:MN}
We consider a set of $d$ discrete random variables $X=\{X_{1},\ldots,X_{d}\}$ where each variable $X_{j}$ takes values from a finite set of outcomes $\mathcal{X}_{j}$. A Markov network over $X$ is a undirected probabilistic graphical model that compactly represents a joint distribution over the variables. The dependence structure over the $d$ variables is specified by an undirected graph $G=(V,E)$ where the nodes $V=\{1,\ldots,d\}$ correspond to the indices of the variables $X$ and the edge set $E\subseteq \{ V\times V\}$ represents dependencies among the variables. We will use the terms node and variable interchangeably throughout this article. The complete set of undirected graphs is denoted by $\mathcal{G}$.

A node $i$ is a neighbor of $j$ (and vice versa) if $\{i,j\}\in E$ and the set of all neighbors of $j$ is called its Markov blanket, which is denoted by $mb(j)=\{ i\in V:\{i,j\}\in E \}$. A clique in a graph is a subset of nodes, $C\subseteq V$, for which every pair of nodes are connected by an edge, that is $\{i,j\}\in E$ if $i,j\in C$. A clique is considered maximal if it cannot be extended by including an additional node without violating the clique criterion. The set of maximal cliques associated with a graph is denoted by $\mathcal{C}(G)$. The variables corresponding to a subset of nodes, $S\subseteq V$, are denoted by $X_{S}=\{X_{j}\}_{j\in S}$ and the corresponding joint outcome space  is specified by the Cartesian product $\mathcal{X}_{S}=\times_{j\in S}\mathcal{X}_{j}$. The cardinality of an outcome space is denoted by $|\mathcal{X}_{S}|$. We use a lowercase letter $x_{S}$ to denote that the variables have been assigned a specific joint outcome in $\mathcal{X}_{S}$. A dataset $\mathbf{x}=(\mathbf{x}_{1},\ldots,\mathbf{x}_{n})$ refers to a collection of $n$ i.i.d. complete joint observations $\mathbf{x}_{k}=(x_{k,1},\ldots,x_{k,d})$ over the $d$ variables, that is $x_{k,j}\in \mathcal{X}_{j}$ for all $k$ and $j$. 

In addition to the graph, to fully specify a Markov network one must also define a probability distribution that satisfies the restrictions imposed by the graph $G$. We restrict the models to positive and faithful distributions unless otherwise mentioned. A distribution is said to be faithful to $G$ if it does not satisfy any additional independencies that are not conveyed by the graph. In this case $G$ can be considered a true representation in the sense that no artificial dependencies are introduced. We use $\theta_{G}$ to denote the set of parameters describing a distribution of a model with graph $G$. The parameter space $\Theta_{G}$ contains all possible instantiations of $\theta_{G}$ corresponding to a distribution satisfying $G$. Finally, we use $p(x_{A}\mid x_{B})$ as an abbreviated notation for the conditional probability $p(X_{A}=x_{A}\mid X_{B}=x_{B})$, while $p(X_{A}\mid X_{B})$ represents the corresponding family of conditional distributions.

The concept of graphical models is based on the assumption of modularity manifested in the factorization of the joint distribution. In particular, the (positive) joint distribution of a Markov network can be factorized over the maximal cliques in the graph according to
\begin{equation}
p(x)=\frac{1}{Z}\overset{}{\underset{{\scriptstyle C \in \mathcal{C}(G)}}{\prod}}\phi(x_{C})\label{eq:cliqueFAC}
\end{equation}
where $\phi (x_C):\mathcal{X}_{C}\rightarrow \mathbb{R}_{+}$ is a clique factor (or potential) and $Z=\sum_{x \in \mathcal{X}} \prod_{C \in \mathcal{C}(G)} \phi(x_{C})$ is a normalizing constant known as the partition function. Markov networks are often also parameterized in terms of a log-linear model in which each clique factor is replaced by an exponentiated weighted sum of features according to
\begin{equation}
p(x)=\frac{1}{Z}\exp\left(\overset{}{\underset{{\scriptstyle f_{K} \in \mathcal{F}}}{\sum}} w_{K} f_{K}(x_{K})\right)\label{eq:cliquePOT}
\end{equation}
where $\mathcal{F}=\{f_{K}\}$ is the set of feature functions and $\mathcal{W}=\{w_{K}\}$ is the corresponding set of weights. A feature function $f_{K}:\mathcal{X}_{K}\rightarrow \mathbb{R}$ maps each value $x_{K}\in\mathcal{X}_{K}$ for some $K\subseteq V$ to a numerical value, typically it is in the form of an indicator function that equals $1$ if the value matches a specific feature and $0$ otherwise. Every Markov network can be encoded as a log-linear model by defining a feature as an indicator function for every assignment of $X_{C}$ for each $C\in\mathcal{C}(G)$. In this case, the weights in \eqref{eq:cliquePOT} correspond to the natural logarithm of the clique factors in \eqref{eq:cliqueFAC}. Conversely, a log-linear model over $X$ implicitly induces the graph of a Markov network by imposing an edge $\{i,j\}$ for every pair of variables appearing in the same domain of some feature function $f_{K}(x_{K})$, that is $\{i,j\}\in E$ if   $\{i,j\}\subseteq K$.   

The absence of edges in the graph $G=(V,E)$ of a Markov network encodes statements of conditional independence. The variables $X_{A}$ are conditionally independent of the variables $X_{B}$ given the variables $X_{S}$ if $p(X_{A}\mid X_{B},X_{S})=p(X_{A}\mid X_{S})$ holds. We denote this by 
\[
X_{A}\perp X_{B}\mid X_{S}.
\] 
The dependence structure of a Markov network can be characterized by the following Markov properties:
\begin{enumerate}
\item Pairwise Markov property: $X_{i}\perp X_{j}\mid X_{V\setminus \{i,j\}}$ for all $\{i,j\}\not\in E$.
\item Local Markov property: $X_{i}\perp X_{V\setminus \{mb(i)\cup i\}}\mid X_{mb(i)}$ for all $i\in V$.
\item Global Markov property: $X_{A}\perp X_{B}\mid X_{S}$ for all disjoint subsets $(A,B,S)$ of $V$ such that $S$ separates $A$ from $B$. 
\end{enumerate}
Although the strength of the above properties differ in general, they are proven to be equivalent under the current assumption of positivity of the joint distribution \citep{Lauritzen1996}. While the last property is sufficient in the sense that it captures the entire set of independencies induced by a network, the first two properties are also useful since they allow one to focus on smaller sets of independencies. In particular, our MPL approach for structure learning is based on the local Markov property.

\section{Structure learning}
There are two main tasks associated with fitting graphical models to data; parameter estimation and structure learning. In this work, we focus entirely on the latter. By structure learning, we refer to the process of deducing the dependence structure from a set of data assumed to be generated from an unknown Markov network. The structure learning problem can be considered a model class learning problem in the sense that each specific structure alone represents a class of models. In many applications, the structure is a goal in itself in the sense that one wants merely to gain a qualitative insight into the dependence structure of an underlying  process. However, given a known structure, the problem of model parameter estimation is simplified. Hence, if the distribution needs also to be explicitly estimated, this can be achieved by using any of the several existing methods conditional on the fixed structure learned by our approach. 

\subsection{Hypothesis space}
When learning the structure of a Markov network the considered space of model classes, or hypothesis space, can be formulated in terms of different degrees of granularity \citep{Koller2009}. The most fine-grained structure learning methods aim at recovering distinct features in the log-linear parameterization \eqref{eq:cliquePOT}, this approach is commonly referred to as feature selection \citep{DellaPietra1997,Lee2006,Hofling2009,Ravikumar2010,Lowd2010}. The advantage of a very detailed structure is that it enables the model to better emulate the properties of a distribution without imposing redundant parameters. One possible drawback of this formulation is the risk of overfitting the structure through long specialized features. Since every pair of variables in a feature results in an edge, such parameterizations can obscure the connection to the graph structure in the sense that sparsity in the number of features does not in general correspond to sparsity in the number of edges in the graph. In contrast to the very specific feature selection problem, the model space of our approach is formulated directly in terms of the graph structure alone and the complexity of a model is defined by the size of the maximal cliques in the network.  

Although dense graphs are not necessarily unfavorable, there are several situations where a sparse graph is preferred. In particular, when the ultimate goal is knowledge discovery, a dense graph may in the worst case hide the primary layer of the dependency pattern. Another important aspect is the feasibility of performing probabilistic inference in the model. One of the main inference tasks for graphical models is the process of computing the posterior probability of a list of query variables given some observed variables. Inference methods designed for this purpose often exploit the sparsity of the graph structure and dense graphs inevitably hamper the efficiency of such algorithms. 

\subsection{Different approaches}
Structure learning methods can roughly be divided into two categories; constraint-based and score-based methods. Constraint-based approaches aim at inferring the structure through a series of independence tests based on the Markov properties, \citep{Spirtes2000,Tsamardinos2003,Bromberg2009,Anandkumar2012}. This approach is appealing in the sense that the independently performed tests can be combined into a structure through a divide-and-conquer approach. Under the assumptions that the distribution is faithful to graph structure and that the tests are correct, the true structure can be reconstructed. However, constraint-based approaches can be quite sensitive to failures in individual tests in the sense that a wrong answer from an independence test can mislead the network construction procedure \citep{Koller2009}. In practice, rather large sample sizes may be required for the independence tests to yield correct answers. 

The score-based approach formulates structure learning as an optimization problem. One defines an objective or score function according to which the plausibility of each candidate in the model space can be evaluated. Since score functions consider the whole structure at once, they can be less sensitive to individual failures. The disadvantage of the score-based approach is that it usually requires use of an optimization algorithm. This poses an obvious problem since the search space for $d$ nodes consists of $2\hat{ \ } {d \choose 2}$ distinct undirected graphs. Finding the global optimum in such enormous combinatorial spaces becomes intractable already for moderate-sized models. For this reason, a selection of heuristic search algorithms have been developed for the sole purpose of finding high-scoring networks and many of them have been shown to work well in practice.  

The most commonly used objective function is the likelihood of the data given a graph, 
\begin{equation*}
l(\theta_{G} ; \mathbf{x})=p(\mathbf{x}\mid \theta_{G})=\prod_{k=1}^{n} p(\mathbf{x}_{k} \mid \theta_{G}),
\end{equation*}
or, in practice, the corresponding log-likelihood function,
\begin{equation*}
\ell(\theta_{G} ; \mathbf{x})=\log l(\theta_{G} ; \mathbf{x}).
\end{equation*}
By maximizing the (log-)likelihood, the model fit to the data is maximized. Although there exists no analytical solution for non-chordal Markov networks, the concavity of the likelihood function enables it to be maximized by numerical optimization for moderate-sized models. The maximum likelihood alone is not an appropriate objective function since it obtains its maximum value under the complete graph due to noise in the data. One option is to constrain the expressiveness of the graphs in the model space, for example by only considering tree structures \citep{Chow1968}. A problem with such a constraint is that it may easily end up limiting the model space to networks not suitable for modeling the data. A more popular approach is to regulate the fit by adding a sparsity-promoting penalty function to the log-likelihood \citep{Akaike1974,Schwarz1978,Lee2006}.

In contrast to the above methods where the complexity of a model is penalized explicitly, the Bayesian framework provides an alternative by implicitly preventing overfitting. In the Bayesian approach a graph is scored by its posterior probability given the data,
\begin{equation}
p(G \mid \mathbf{x})=\frac{p(\mathbf{x}\mid G) \cdot p(G)}{p(\mathbf{x})}.\label{eq:PostProb}
\end{equation}
In practice it suffices to consider the unnormalized posterior probability 
\begin{equation}
p(G , \mathbf{x})=p(\mathbf{x}\mid G) \cdot p(G),\label{eq:unnPostProb}
\end{equation}
since $p(\mathbf{x})$ is a normalizing constant that can be ignored when comparing graphs. The key factor of \eqref{eq:unnPostProb} is $p(\mathbf{x}\mid G)$ which is the marginal likelihood (ML) of the data given the network structure (also called the evidence). To evaluate the ML, one must integrate the likelihood function over all parameter values satisfying the restrictions imposed by the graph according to 
\begin{equation}
p(\mathbf{x}\mid G)=\int_{\Theta_{G}}l(\theta_{G} ; \mathbf{x})\cdot f(\theta_{G})d\theta_{G}, \label{eq:margLL}
\end{equation}
where $f(\theta_{G})$ is a prior distribution that assigns a weight to each $\theta_{G} \in \Theta_{G}$. Since the ML accounts for the parameter uncertainty through the prior, it implicitly regulates the fit to the data against the complexity of the network.

A drawback of the ML is that it is extremely hard to evaluate for non-chordal Markov networks. For this reason various penalized maximum likelihood objectives have naturally been preferred. In particular, \citet{Schwarz1978} introduced the Bayesian information criterion (BIC) as an asymptotic approximation of the ML. Still, due to the partition function, even maximum likelihood based techniques become intractable for larger models and require use of approximate inference. Therefore, in the next section we derive an alternative Bayesian-type score applicable also to very large systems.

Given a scoring function, it is still necessary to specify a search algorithm to find high-scoring networks since the discrete search space is in general too large for an exhaustive evaluation. To avoid the discrete nature of the model space, \cite{Lee2006} introduced an $L_{1}$-based penalty to reformulate the structure learning problem as a convex optimization problem over the continuous parameter space. This is an elegant technique that has been further developed \citep{Hofling2009,Ravikumar2010} for the special class of binary pairwise Markov networks for which the method is especially well-suited. Each edge in such a network is associated with a single parameter and forcing an edge parameter to zero is equivalent to removing the corresponding edge from the network. In fact, in this case the problem formulation of feature selection and graph structure discovery are equivalent due to the one-to-one correspondence between edges and features. For more general networks, sparsity must be enforced to groups of parameters in order to achieve sparsity in the number of edges in the resulting graph \citep{Schmidt2010}. For the direct approach of \citet{Lee2006} the issue of maximizing the likelihood function still remains. 

The main difference between constraint- and score-based methods is the level at which they approach the problem \citep{Koller2009}. Score-based methods works on a global level by considering the whole structure at once. This makes them less sensitive to individual failures but has a negative effect on their scalability. The local approach of constraint-based methods allows them to scale up well but it makes them more sensitive to failures in the individual tests. Although the MPL as such would fall into the score-based category, under the optimization strategy introduced in Section \ref{sec:Search}, our MPL method is rather a hybrid by which we aim to achieve scalability as well as reliable performance.  

\section{Marginal pseudo-likelihood}\label{sec:MPL}
In order to avoid problems associated with the evaluation of the true likelihood function, one can preferably use alternative objectives that possess favorable properties from a computational perspective. In this work we consider the commonly used pseudo-likelihood, originally introduced by \citet{Besag1972}, from a Bayesian perspective.
\subsection{Derivation}
The pseudo-likelihood function approximates the likelihood function by a factorization into conditional likelihood functions according to
\begin{equation*}
pl(\theta ; \mathbf{x})=\prod_{k=1}^{n}\prod_{j=1}^{d} p(x_{k,j} \mid x_{k,V\setminus j},\theta).
\end{equation*} 
For a fixed graph structure $G$, the local Markov property implies that a variable in a Markov network is independent of the remaining variables given its Markov blanket such that 
\[
p(X_{j}\mid X_{V\setminus j},G)=p(X_{j}\mid X_{mb(j)},G)
\]
must hold. Consequently, the pseudo-likelihood for a fixed graph is given by
\begin{equation}
pl(\theta_{G} ; \mathbf{x})=\prod_{k=1}^{n} \prod_{j=1}^{d} p(x_{k,j} \mid x_{k,mb(j)},\theta_{G}).\label{eq:PLforG}
\end{equation}
In terms of the log-linear parameterization \eqref{eq:cliquePOT}, the pseudo-likelihood approximation offers huge computational savings compared to the true likelihood since the global normalizing constant in the likelihood function is replaced by $d$ local normalizing constants. By replacing the likelihood with the pseudo-likelihood, methods originally based on the maximum likelihood have been extended to work on larger systems. For example, the pseudo-likelihood approximation of \citet{Hofling2009} and the closely related method by \citet{Ravikumar2010} highlight how the original idea of \citet{Lee2006} can be extended to higher dimensions. \citet{Ji1996} and \citet{Csiszar2006} both derived a pseudo-likelihood version of the Bayesian information criterion by \citet{Schwarz1978}. An encouraging aspect is that several pseudo-likelihood approaches have been shown to enjoy consistency under the assumption that the data is generated by a distribution in the model class \citep{Ji1996,Csiszar2006,Ravikumar2010}. 

From a Bayesian perspective, the structural form of \eqref{eq:PLforG} offers an interesting possibility. In fact, under certain assumptions it enables an analytical evaluation of the integral
\begin{equation}
\hat p(\mathbf{x}\mid G)=\int_{\Theta_{G}} pl(\theta_{G} ; \mathbf{x})\cdot f(\theta_{G})d\theta_{G} \label{eq:PLint}
\end{equation}
which is here referred to as the marginal pseudo-likelihood (MPL). We parameterize the conditional probabilities associated with the pseudo-likelihood function of a graph by 
\begin{equation}\label{eq:MPLpara}
\theta_{ijl}=p(X_{j}=   x_{j}^{(i)}\mid X_{mb(j)}=x_{mb(j)}^{(l)}) \text{ where } \theta_{ijl}>0 \text{ and } \sum_{i=1}^{r_j}\theta_{ijl}=1.
\end{equation}
The indices $i=1,\ldots,r_{j}$ and $l=1,\ldots,q_{j}$, where $r_{j}=|\mathcal{X}_j|$ and $q_{j}=|\mathcal{X}_{mb(j)}|=\prod_{i\in mb(j)}r_i$, represent the configurations of the variable and its respective Markov blanket. The above set of graph-specific parameters is by no means a compact representation of a Markov network, in fact, it is a quite crude over-parameterization. Rather than actual model parameters, they should be considered temporary nuisance parameters, used solely for computational convenience, in solving the structure learning problem. Similarly as above, we denote the counts of the corresponding configurations in $\mathbf{x}$ by
\[
n_{ijl}=\sum_{k=1}^{n} \mathbf{I}\left[(x_{k,j},x_{k,mb(j)})=(x_{j}^{(i)},x_{mb(j)}^{(l)})\right] \text{ and } n_{jl}=\sum_{i=1}^{r_j} n_{ijl}.
\]
The pseudo-likelihood function can now be expressed in terms of our above notation by
\begin{equation}
pl(\theta_{G} ; \mathbf{x})=\prod_{j=1}^{d} \prod_{l=1}^{q_{j}} \prod_{i=1}^{r_{j}} \theta_{ijl}^{n_{ijl}}.\label{eq:PLfac}
\end{equation}
Under the current parameterization it is easy to make out certain structural similarities between the above pseudo-likelihood function and the likelihood function of a Bayesian network under a standard conditional parameterization \citep[see e.g.][]{Koller2009}. In a Bayesian network the $l$-index would be associated with configurations of parent sets instead of configurations of Markov blankets. The parent sets must be such that they satisfy the acyclicity constraint imposed by a DAG whereas the Markov blankets must be mutually consistent. Under certain assumptions listed by \citet{Heckerman1995} the ML of a Bayesian network has a nice analytical expression that factorizes variable-wise making it attractive for the task of structure learning. Using a corresponding set of assumptions we would like to achieve something similar for the ML. We consider the parameters defined in \eqref{eq:MPLpara} in terms of the sets
\[
\theta_{jl}=\cup_{i=1}^{r_j} \{ \theta_{ijl} \}\text{, }\theta_{j}=\cup_{l=1}^{q_j} \{ \theta_{jl} \} \text{, and }\theta_{G}=\cup_{j=1}^{d} \{ \theta_{j} \}.
\]
One of the fundamental assumptions behind the ML for Bayesian networks is an assumption regarding global and local parameter independence \citep[Assumption 2,][]{Heckerman1995}. This assumption ultimately justifies a factorization of the parameter prior. We need to factorize the parameter prior in \eqref{eq:PLint} in a corresponding fashion according to  
\[
f(\theta_{G}) =\prod_{j=1}^{d}f(\theta_{j}) = \prod_{j=1}^{d}\prod_{l=1}^{q_j}f(\theta_{jl}),
\]
implying that $\theta_{j} \perp \theta_{j'}$ for $j\not= j'$ (global parameter independence) and $\theta_{jl} \perp \theta_{jl'}$ for $l\not= l'$ (local parameter independence). Whereas parameter independence can be a quite reasonable assumption in a Bayesian network parameterization, in our case it directly violates the properties of a Markov network. The conditional distributions, represented by our parameters, are connected to each other in the sense that they must satisfy certain algebraic relations for them to be consistent with a Markov network. We do not elaborate on these relations but we note that they directly translate to restrictions between the corresponding parameter sets. At this point, the parameter independence assumption is mainly justified by the induced computational savings. In Section \ref{sec:related} we discuss the implications of the assumption more in detail from another perspective. Another fundamental assumption, necessary for our derivation, is to restrict each parameter set $\theta_{jl}$ to follow a Dirichlet distribution
\[
\theta_{jl} \sim \text{Dirichlet}(\alpha_{1jl},\ldots,\alpha_{r_{j}jl}),
\]
where $\alpha_{1jl},\ldots,\alpha_{r_{j}jl}$ are hyperparameters for which we denote $\alpha_{jl}=\sum_{i=1}^{r_j} \alpha_{ijl}$. 

Under the established assumptions, the integral in \eqref{eq:PLint} can be reordered into a product of local integrals. Since the Dirichlet distribution is a conjugate prior of the multinomial distribution, each local integral is easily solved using standard Bayesian calculations:
\begin{equation*}
\begin{aligned}
\hat p(\mathbf{x}\mid G) 	& = \int_{\Theta_{G}} pl(\theta_{G} ; \mathbf{x})\cdot f(\theta_{G})d\theta_{G} \\
			         	& = \prod_{j=1}^{d} \prod_{l=1}^{q_{j}} \int_{\Theta_{jl}} \prod_{i=1}^{r_{j}} \theta_{ijl}^{n_{ijl}}\cdot f(\theta_{jl})d\theta_{jl} \\
				& = \prod_{j=1}^{d} \prod_{l=1}^{q_{j}} \frac{\Gamma ( \alpha_{jl})}{\Gamma ( n_{jl}+\alpha_{jl})} \prod_{i=1}^{r_{j}} \frac{\Gamma ( n_{ijl}+\alpha_{ijl})}{\Gamma ( \alpha_{ijl})}
\end{aligned}
\end{equation*}
In practice, the logarithm of the formula is used since it is computationally more manageable.
  
To evaluate the above expression it is necessary to define the hyperparameters. We want to specify a symmetric prior since we assume that there is no prior knowledge favoring one parameter in $\{\theta_{1jl},\ldots,\theta_{r_{j}jl}\}$ over any of the others. We achieve this by modifying a prior originally defined for Bayesian networks by \citet{Buntine1991} such that the hyperparameters are determined according to
\[
\alpha_{ijl}=\frac{N}{|\mathcal{X}_{j}| \cdot |\mathcal{X}_{mb(j)}|}=\frac{N}{r_{j} \cdot q_{j}},
\]  
where $N$ is the equivalent sample size adjusting the strength of the prior.

\subsection{Properties\label{sec:Properties}}
The MPL possesses several advantageous properties. The parameter prior offers a natural regularization that prevents overfitting. Methods that explicitly penalize the degree of regularization are sensitive to the choice of some tuning parameter, which usually has to be determined empirically. In contrast, the MPL requires specification of the hyperparameters in the Dirichlet distribution. In our formulation this boils down to setting a value on the equivalent sample size $N$. \citet{Silander2007} show that the maximum a posteriori (MAP) Bayesian network structure optimization problem is indeed sensitive to the choice of value on the equivalent sample size parameter. Due to the similarity between the MPL and the BDeu score considered in \citet{Silander2007}, one would expect the MPL to display a similar behavior. In this work we primarily focus on the setting where $N=1$, and simulations are used to demonstrate the adequacy of this choice.

An important property preferably satisfied by a scoring function is consistency. By consistency we mean that, under the assumption that the generating distribution is faithful to a Markov network structure, the score will favor the true graph when the sample size tends to infinity. The following theorem establishes that MPL is indeed a consistent scoring function for Markov networks. 
\begin{theorem}\label{thm:MPLconsistency}
Let $G^{*} \in\mathcal{G}$ be the true graph structure, of a Markov network over $(X_{1},\ldots,X_{d})$, with the corresponding Markov blankets $mb(G^{*})=\{ mb^{*}(1),\ldots,mb^{*}(d) \}$. Let $\theta_{G^{*}}\in\Theta_{G^{*}}$ define the corresponding joint distribution which is faithful to $G^{*}$ and from which a sample $\mathbf{x}$ of size $n$ is obtained. The local MPL estimator
\begin{equation*}
\skew{6}\widehat{m}b(j) = \underset{mb(j)\subseteq V\setminus j}{\arg \max} \ p(\mathbf{x}_{j} \mid \mathbf{x}_{mb(j)})
\end{equation*} 
is consistent in the sense that $\skew{6}\widehat{m}b(j) = mb^{*}(j)$ eventually almost surely as $n\to \infty$ for $j=1,\ldots,d$. Consequently, the global MPL estimator
\begin{equation*}
\hat G = \underset{G\in\mathcal{G}}{\arg \max} \ \hat p(\mathbf{x}\mid G)
\end{equation*} 
is consistent in the sense that $\hat G = G^{*}$ eventually almost surely as $n\rightarrow \infty$.
\end{theorem}
{\bf Proof } See Appendix A.

Although consistency alone is a reassuring theoretical property, it is also important to recognize its limitations. In particular, the assumptions under which the result is obtained rarely hold in practice. Therefore it is important to investigate how well the MPL performs in practice. In Section \ref{sec:Results} we do a series of large-scale numerical simulations to investigate how well the MPL performs in combination with the search algorithms introduced in Section \ref{sec:Search}. Before that, we conduct a small-scale simulation study to gain an insight into the behavior of the MPL both when it comes to choosing the optimal graph as well as ranking the most plausible graphs. 

In the first part of the experiment we restrict the model space to chordal graphs. By doing so we can calculate the ML of each considered graph and compare it to the MPL. The ML of a chordal graph is usually calculated by factorizing the likelihood according to the maximal cliques and separators of the graph \citep[see e.g.][]{Corander2008}, however, there is an alternative approach. To any chordal graph, there exists a collection of Markov equivalent DAGs, each encoding an equivalent dependence structure as the undirected graph. We can therefore evaluate the ML of an undirected chordal graph by the BDeu metric \citep{Buntine1991} of one of the equivalent DAGs. Since the BDeu metric assigns the same score to all Markov equivalent DAGs, it does not depend on which DAG being picked for evaluation. The main advantage of the DAG-based approach is that the structural form of the ML is similar to the MPL since the factorization of the likelihood and the choice of hyperparameters are done analogously. Consequently, we can apply both methods under fairly similar conditions such that the different behaviors are primarily due to different fundamental characteristics of the score functions. 


\begin{figure}[tb]
\begin{subfigure}{0.32\textwidth}
\captionsetup{skip=-5pt}
\begin{center}
\includegraphics[width=3.5cm]{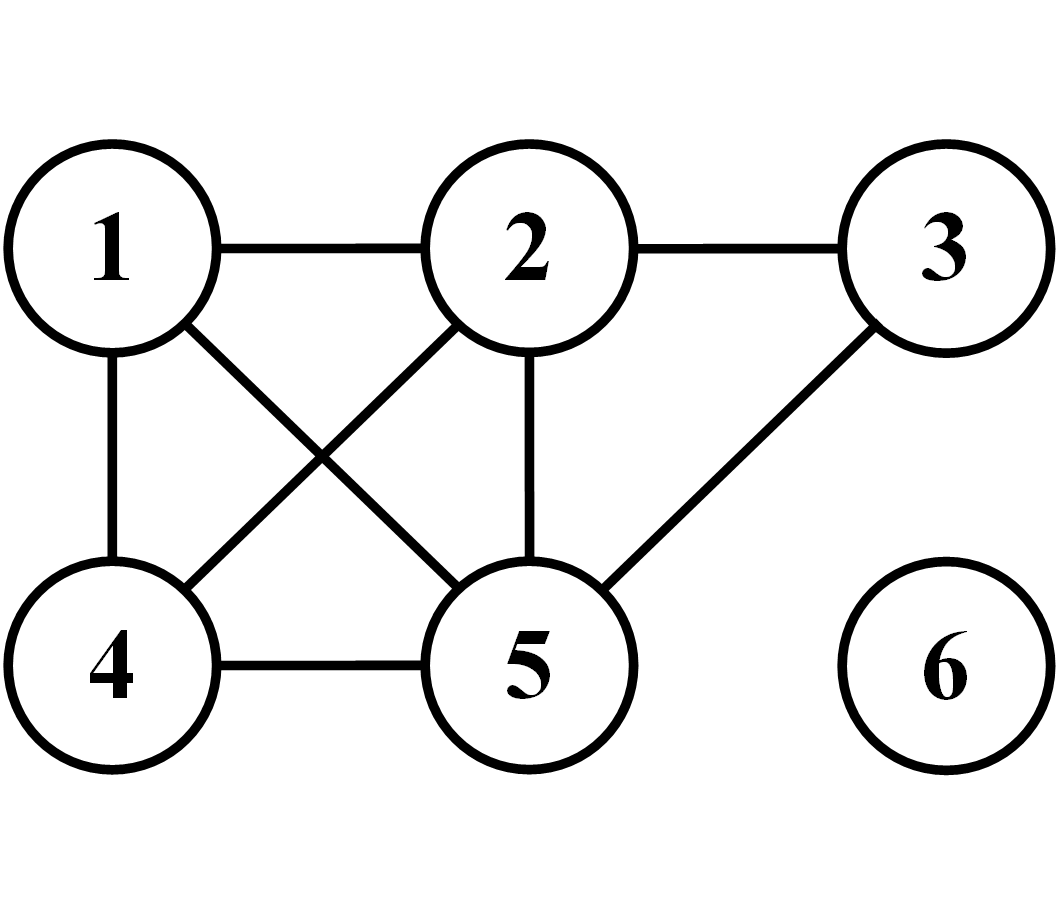}
\end{center}
\caption{\label{fig:graph_1}}
\end{subfigure}
\begin{subfigure}{0.32\textwidth}
\captionsetup{skip=-5pt}
\begin{center}
\includegraphics[width=3.5cm]{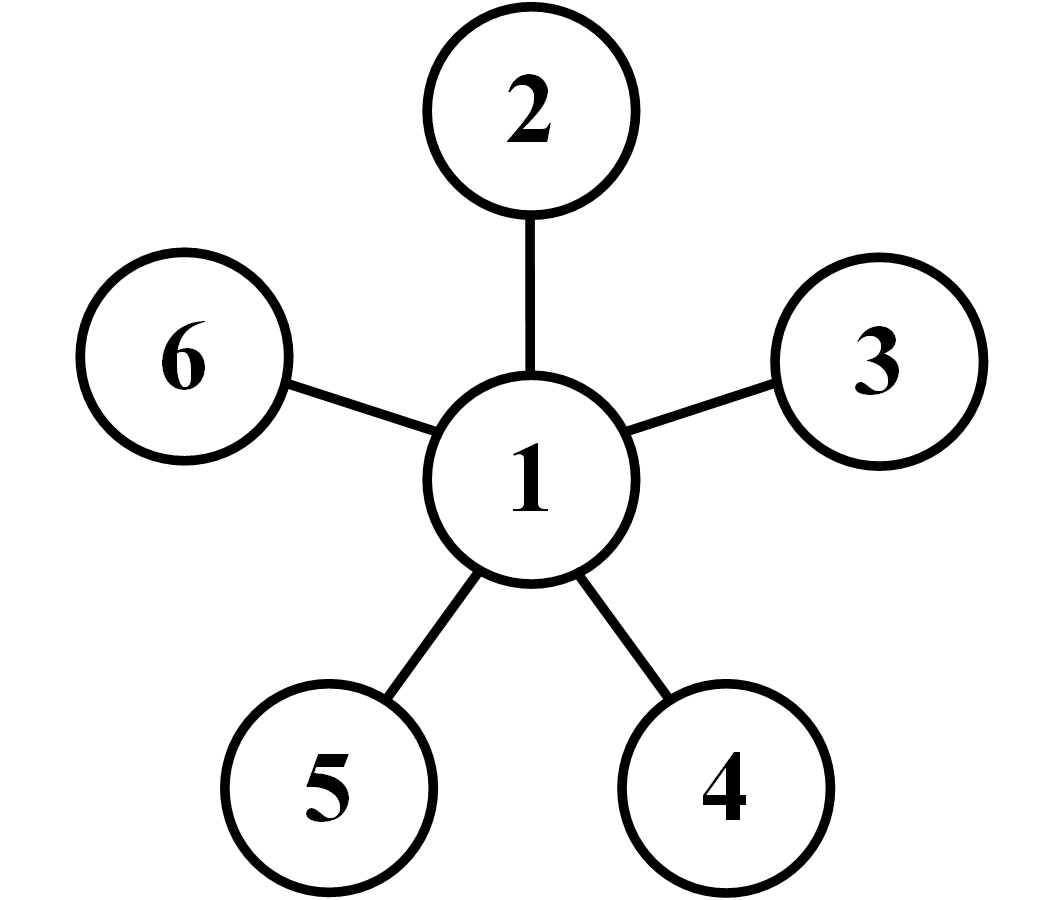}
\end{center}
\caption{\label{fig:graph_2}}
\end{subfigure}
\begin{subfigure}{0.32\textwidth}
\captionsetup{skip=-5pt}
\begin{center}
\includegraphics[width=3.5cm]{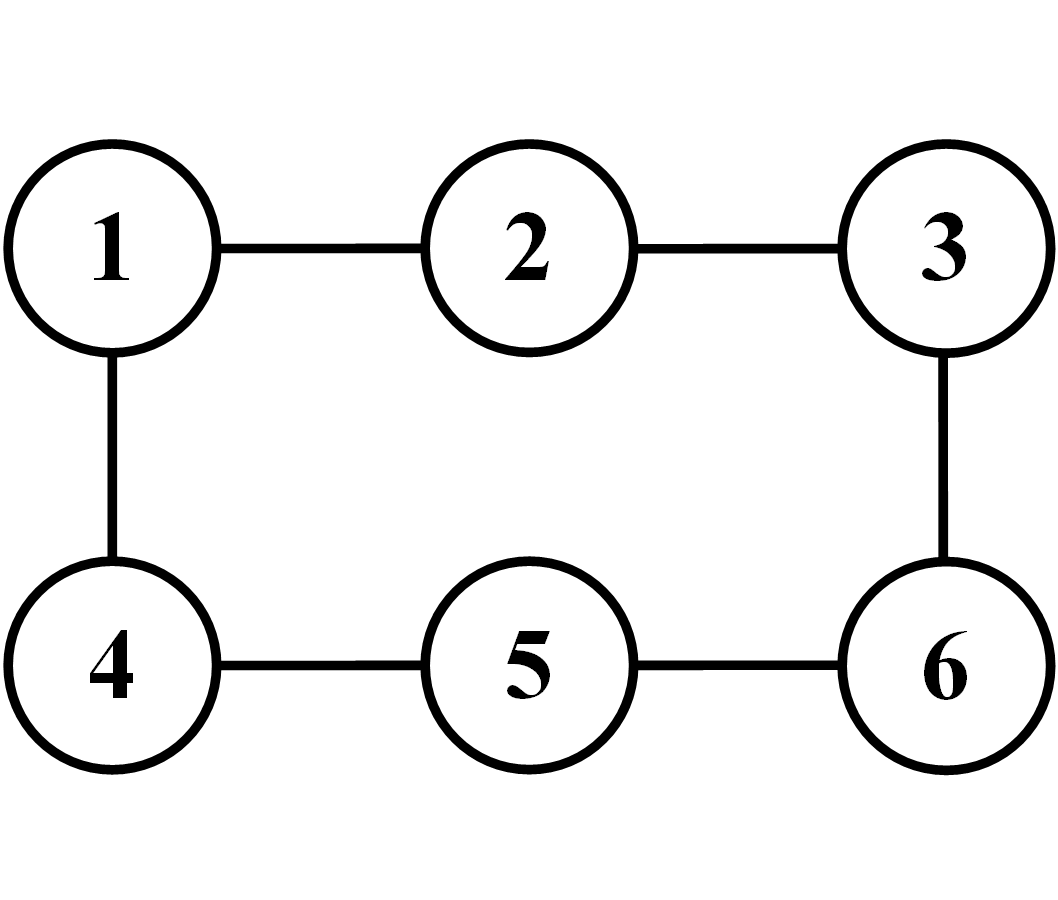}
\end{center}
\caption{\label{fig:graph_3}}
\end{subfigure}
\caption{Graphs used in the simulations in Section \ref{sec:Properties}.\label{fig:graphs}}
\end{figure} 
First, we used the graph in Figure \ref{fig:graph_1} as base for the generating model. The number of possible graphs over six nodes is $2\hat{ \ } {6 \choose 2}=32768$ and $18154$ of these are chordal. To generate a distribution according to a graph, we assigned values to the maximal clique factors in \eqref{eq:cliqueFAC} by independently sampling from a uniform distribution over $(0,1)$. We generated ten distributions and for each distribution we generated ten samples. The final results are thus averaged over hundred samples. We performed an exhaustive evaluation of the chordal graphs and listed the hundred highest ranked graphs for the respective score.

\begin{figure}[tb]
\begin{subfigure}{0.5\textwidth}
\captionsetup{skip=-5pt}
\begin{center}
\includegraphics{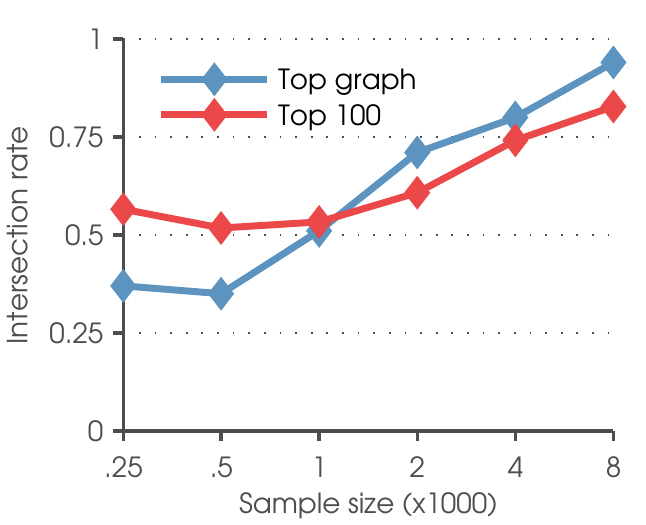}
\end{center}
\caption{\label{fig:Ex1a}}
\end{subfigure}
\begin{subfigure}{0.5\textwidth}
\captionsetup{skip=-5pt}
\begin{center}
\includegraphics{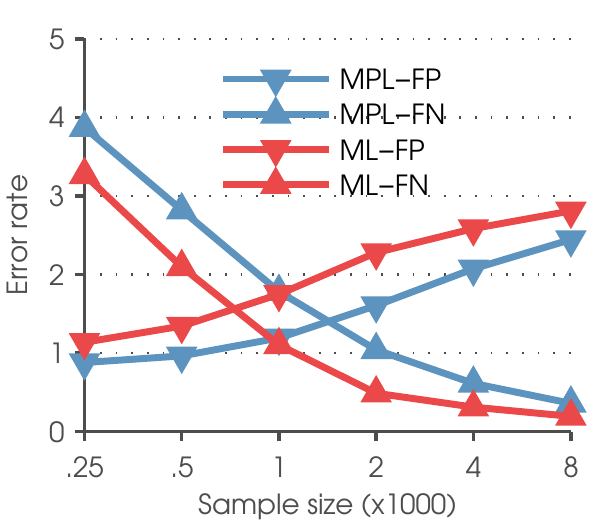}
\end{center}
\caption{\label{fig:Ex1b}}
\end{subfigure}
\caption{Comparison of the MPL and ML graph rankings for different sample sizes. In (a) the similarity of rankings are compared for the top ranked graph and the 100 top ranked graphs. In (b) the average FP and FN rates are compared for the 100 top ranked graphs.}
\end{figure} 
To begin with, we consider the similarity of the rankings. Figure \ref{fig:Ex1a} illustrates the rate at which the ML and MPL scores agree on the top graph as well the percentage of graphs included in both of the top 100 rankings. With an increased sample size, the two scores show an increased conformity in how they rank the graphs. To further investigate the differences, Figure \ref{fig:Ex1b} illustrates the average rate of falsely added edges (False Positives, FPs) and falsely omitted edges (False Negatives, FNs) among the 100 top ranked graphs. Although the overall error rates are quite similar, there is a key difference between the MPL and the ML which is nicely illustrated by the figure.  Since the MPL in a sense over-determines the dependence structure, it is more conservative in terms of adding edges. This phenomenon is clearly reflected by the MPL consistently having a lower false positive rate and a higher false negative rate than the ML. The difference becomes less distinct for larger sample sizes which is in concordance with Figure \ref{fig:Ex1a}.

One drawback of the above mentioned characteristic is that it makes the MPL less sample efficient than the ML in terms of identifying the correct graph. 
\begin{figure}[tb]
\begin{subfigure}{0.33\textwidth}
\captionsetup{skip=-5pt}
\begin{center}
\includegraphics{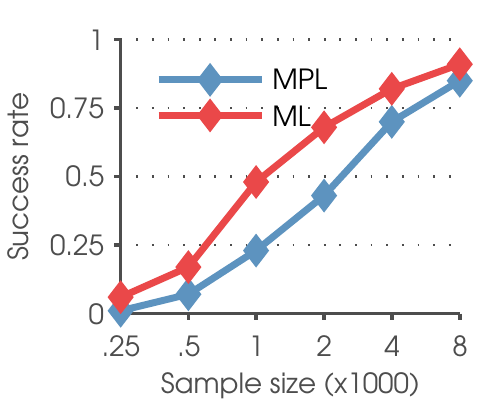}
\end{center}
\caption{\label{fig:Ex2a}}
\end{subfigure}
\hspace{-0.20cm}
\begin{subfigure}{0.33\textwidth}
\captionsetup{skip=-5pt}
\begin{center}
\includegraphics{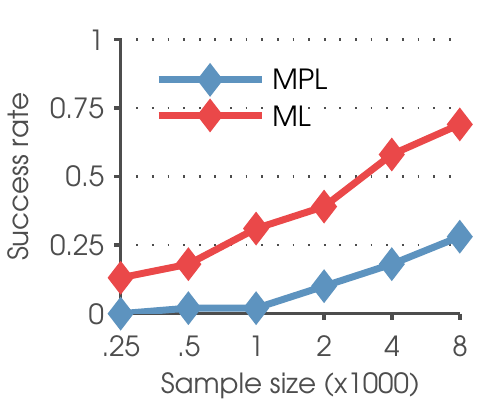}
\end{center}
\caption{\label{fig:Ex2b}}
\end{subfigure}
\hspace{-0.20cm}
\begin{subfigure}{0.33\textwidth}
\captionsetup{skip=-5pt}
\begin{center}
\includegraphics{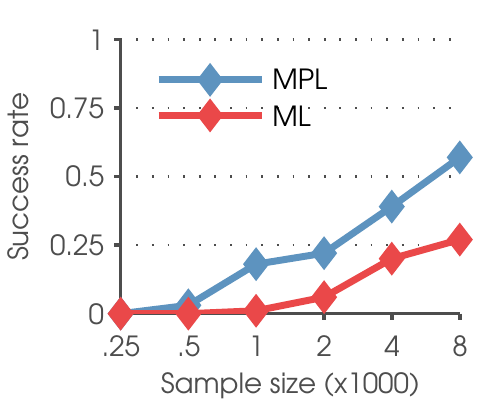}
\end{center}
\caption{\label{fig:Ex2c}}
\end{subfigure}
\caption{Comparison of the MPL and ML top ranked graph for the graphs in Figure \ref{fig:graphs}. The success rate, which is plotted against the sample size, refers to the rate at which the correct graph is identified except for in (c) where the ML success rate refers to the rate at which the top ranked graph contains all the true edges.}
\end{figure} 
In Figure \ref{fig:Ex2a} we have plotted the rate at which the true graph was ranked as optimal by the respective score. Although the curves eventually converge for large enough sample sizes, the ML outperforms the MPL for all the considered sample sizes. This weakness is exaggerated for graphs containing large Markov blankets compared to the maximal clique sizes. As an ultimate example of this, consider the star graph in Figure \ref{fig:graph_2} for which there is one hub node connected to all the other nodes. In Figure \ref{fig:Ex2b} we see that the ML has a clear advantage over the MPL, for this type of graph, for limited sample sizes. Still, the curves will eventually converge as confirmed by Theorem \ref{thm:MPLconsistency}.

As expected, the ML is to be preferred over the MPL when it comes to picking the optimal chordal graph. However, we conclude this section by giving an example that illustrates the importance of going beyond chordal graphs, which for larger systems only make up a small fraction of the graph space. In the last experiment we also consider non-chordal graphs. In particular, we based our generating model on the non-chordal graph in Figure \ref{fig:graph_3}. Since the ML can only be evaluated for chordal graphs, it cannot discover the true graph. Therefore, we change the criterion for success for the ML by looking at the true positives. If the top ranked graph contains all of the true edges, we consider it to be correct. In this setup the MPL clearly outperforms the ML as seen in Figure \ref{fig:Ex2c}. This is a natural consequence considering that the ML needs to add three spurious edges in order to form a chordal graph that contains all the true edges. As in the previous case, the curves will eventually converge for large enough sample sizes. Still, a model based on a graph with spurious edges contains redundant parameters which inevitably destabilize a subsequent parameter estimation process.    

\subsection{Computational complexity\label{sec:MPLcomplex}}
Whereas the computational complexity of the ML is determined by the size of the maximal cliques, the computational complexity of the MPL is determined by the size of the Markov blankets. The (log-)MPL is calculated by the sum
\begin{equation*}
\begin{aligned}
\log \hat p(\mathbf{x}\mid G) 	&=\sum_{j=1}^{d} \log p(\mathbf{x}_{j} \mid \mathbf{x}_{mb(j)},G)\\
					&=\sum_{j=1}^{d} \sum_{l=1}^{q_{j}} \log p(\mathbf{x}_{j} \mid \mathbf{x}_{mb(j)}^{(l)},G)\\
					&=\sum_{j=1}^{d} \sum_{l=1}^{q_{j}} \left[ \log \Gamma ( \alpha_{jl})-\log \Gamma ( n_{jl}+\alpha_{jl})+ \sum_{i=1}^{r_{j}}\left[ \log \Gamma ( n_{ijl}+\alpha_{ijl})- \log \Gamma ( \alpha_{ijl})\right] \right],
\end{aligned}
\end{equation*}
which consists of $\sum_{j=1}^{d} q_{j}(2+2r_{j})$ terms. Since $r_{j}=|\mathcal{X}_{j}|$ does not depend on the graph, the number of terms, associated with a node $j$, is mainly determined by the number of Markov blanket configurations, $q_{j}$, which grows exponentially with the size of the Markov blanket. Thereby, the complexity of calculating the MPL of a graph is to a high extent determined by the maximal Markov blanket size. Still, it is important to note that the partial sum
\begin{equation*}
\log p(\mathbf{x}_{j} \mid \mathbf{x}_{mb(j)}^{(l)},G)=\log \Gamma ( \alpha_{jl})-\log \Gamma ( n_{jl}+\alpha_{jl})+ \sum_{i=1}^{r_{j}}\left[ \log \Gamma ( n_{ijl}+\alpha_{ijl})- \log \Gamma ( \alpha_{ijl})\right],
\end{equation*}
does not contribute to the MPL if the corresponding Markov blanket configuration is not represented in the data. Consequently, the maximum number of terms evaluated by a non-naive implementation is $\sum_{j=1}^{d} \min(q_{j},n)(2+2r_{j})$ where $n$ is the number of observations in the dataset. Furthermore, for a large Markov blanket of node $j$, the number of distinct configurations present in a dataset is, in practice, usually far less than $\min(q_{j},n)$.  

If we look at the MPL from an optimization perspective, it is easy to see that its variable-wise decomposition makes it a convenient candidate for search algorithms based on local changes. To compare the plausibility of two graphs, $G_{1}=(V,E_{1})$ and $G_{2}=(V,E_{2})$, we can calculate the ratio of their MPLs,
\begin{equation*}
K(G_{1},G_{2})=\frac{\hat p(\mathbf{x}\mid G_{1})}{\hat p(\mathbf{x}\mid G_{2})}, 
\end{equation*}
or equivalently the log-ratio,
\begin{equation*}
\log K(G_{1},G_{2})=\log \hat p(\mathbf{x}\mid G_{1})-\log \hat p(\mathbf{x}\mid G_{2}), 
\end{equation*}
which is basically the pseudo-version of log-Bayes factor or the log-Bayes pseudo-factor. Assume there is a single edge difference, 
\[
\{ E_{1}\cup E_{2} \} \setminus \{ E_{1}\cap E_{2}\}=\{i,j\},
\]
between the graphs. This implies that $mb(i)$ and $mb(j)$ are the only Markov blankets that differ in the two graphs. Consequently, log-Bayes pseudo-factor is simply evaluated by
\begin{equation*}
\begin{aligned}
\log K(G_{1},G_{2})=	&\log p(\mathbf{x}_{i} \mid \mathbf{x}_{mb(i)},G_{1})+\log p(\mathbf{x}_{j} \mid \mathbf{x}_{mb(j)},G_{1})-\\
				&\log p(\mathbf{x}_{i} \mid \mathbf{x}_{mb(i)},G_{2})-\log p(\mathbf{x}_{j} \mid \mathbf{x}_{mb(j)},G_{2}) 
\end{aligned}
\end{equation*}
since the rest of the terms cancel each other out.
\subsection{Related work\label{sec:related}}
In addition to the asymptotically equivalent PIC (see proof of Theorem \ref{thm:MPLconsistency}) by \citet{Csiszar2006}, the MPL is very closely related to a class of models known as dependency networks \citep{Heckerman2001}. In fact, the general concept of using pseudo-likelihood for Markov networks has an alternative interpretation in terms of this class of models. 

The distribution of a dependency network is, like a Bayesian network, represented by variable-wise conditional distributions in a pseudo-likelihood type manner. The directed graph of a dependency network may thus, in contrast to a Bayesian network, contain cycles. A dependency network does not in general represent a consistent distribution in the sense that the local distribution cannot be inferred from a joint distribution over all the variables. Consequently, one must rely on Gibbs sampling to perform inference in such models. In contrast to dependency networks, Markov networks always represent a consistent distribution. Still, any Markov network with the undirected graph $G$ can be represented by a consistent dependency network with a symmetric directed graph containing the same structural adjacencies as $G$ \citep[Theorem 1 \& 4][]{Heckerman2001}.

The obvious advantage of dependency networks in terms of structure learning is that the local structure of each node can be learned independently of the rest of the network. The local structures can be inferred using a variety of regression-based techniques. In particular, \citet{Heckerman2001} model the local structures using probabilistic decision trees in conjunction with a Bayesian score originally derived by \citet{Friedman1996} for the purpose of including context-specific independence in the learning process of Bayesian networks. \citet{Lowd2010} converted this approach into a feature selection method by transforming the trees into features of a Markov network. The authors mention the risk of overfitting by generating long specialized features. Due to the feature-edge relation described in Section \ref{sec:MN}, the implication of such overfitting would be emphasized in terms of graph structure discovery. 

The logistic regression approach of \citet{Ravikumar2010} is another method that has a natural interpretation in terms of the dependency network framework. The solutions of the separate regression problems represent a structure of a general dependency network and must be made symmetric in order to be consistent with a structure of a Markov network. In contrast, the problem formulation in the closely related approach by \citet{Hofling2009} ensures that the network is kept symmetric and even consistent during the optimization process.    

Our MPL can be interpreted as the ML of a symmetric dependency network under the standard conditional parameterization defined in \eqref{eq:MPLpara}. Without the parameter independence assumption, the MPL would correspond to the ML of a consistent dependency network under our parameterization. This would make up a very natural option for objective function if it could be evaluated efficiently. Still, the goal of the MPL is to learn a graph structure rather than a specific model. If merely considering the dependence structure, each graph of a Markov network has an equivalent counterpart in terms of a symmetric dependency network structure. Therefore it makes sense to enforce consistency among the Markov blankets, that is, to only consider symmetric dependency networks. In contrast to a general dependency network, the MPL evaluates the inclusion of an edge $\{i,j\}$ by comparing the potential benefit from adding $j$ to $mb(i)$ against the potential loss of adding $i$ to $mb(j)$. 

\section{MPL optimization}\label{sec:Search}
The straightforward global MPL-based optimization problem is formulated by
\begin{equation}
\underset{G\in\mathcal{G}}{\arg \max} \ \log \hat p(\mathbf{x}\mid G)+ \log p(G)\label{eq:argmaxMPL}
\end{equation}  
where $p(G)$ is the graph prior distribution which can account for any prior belief regarding for example the degree of sparsity. To maintain the useful structure of the MPL, the prior must follow a similar decomposition. This is achieved by defining the prior in terms of mutually independent prior beliefs on the individual Markov blankets. In Section \ref{sec:GenEX} we give an example of such a prior, however, in the remainder of the section we assume a uniform prior and the term $p(G)$ is therefore omitted. Still, the methods presented in this section are also directly applicable under any prior that follows the same decomposition as the MPL. 

Due to rapidly growing size of the discrete optimization space, the global optimization problem \eqref{eq:argmaxMPL} is clearly intractable already for moderate-sized systems. Hence, we need to construct an algorithm that finds approximate solutions of satisfactory quality in a reasonable time. To ensure applicability in a genuinely high-dimensional setting, the algorithm is designed to exploit the structural decomposition of the MPL by breaking down the problem into two steps instead of directly approaching the global problem \eqref{eq:argmaxMPL}. 

Since each graph $G$ is uniquely specified by its collection of Markov blankets $mb(G)=\{ mb(j)\}_{j=1}^{d}$, we can reformulate \eqref{eq:argmaxMPL} as
\begin{equation}
\begin{gathered}
\underset{mb(G)\in \times_{j\in V}\mathcal{P}(V\setminus j)}{\arg \max} \ \sum_{j=1}^{d} \ \log p(\mathbf{x}_{j}\mid \mathbf{x}_{mb(j)})\vspace{0.25cm}\\
\text{subject to \ \ }  i\in mb(j) \Rightarrow j \in mb(i) \text{ for all } i,j\in V
\end{gathered}
\label{eq:argmaxMPLrf}
\end{equation}
where $\mathcal{P}(V\setminus j)$ is the power set of $V\setminus j$ representing all possible Markov blankets of node $j$. From \eqref{eq:argmaxMPLrf} it is easy to see that our problem is basically made up of $d$ dependent subproblems that are connected through the consistency constraint. By omitting the constraint we remove the dependence among the subproblems and obtain the relaxed problem
\begin{equation}
\underset{mb(G)\in \times_{j\in V}\mathcal{P}(V\setminus j)}{\arg \max} \ \sum_{j=1}^{d} \ \log p(\mathbf{x}_{j}\mid \mathbf{x}_{mb(j)}).\label{eq:argmaxMPLrelax}
\end{equation}
Since the $d$ subproblems now are independent of each other, we can finally reformulate \eqref{eq:argmaxMPLrelax} by breaking it down into a collection of stand-alone Markov blanket discovery problems, 
\begin{equation}
\underset{{mb}(j)\subseteq V\setminus j}{\arg \max} \ \log p(\mathbf{x}_{j}\mid \mathbf{x}_{mb(j)}) \ \ \ \text{ for } j=1,\ldots,d,\label{eq:argmaxMPLrelaxRF}
\end{equation}
which can be solved completely in parallel considerably improving real time efficiency. Since each individual subproblem in itself is still intractable, in Section \ref{sec:MB_disc} we introduce an efficient deterministic search algorithm that gives an approximate solution.  

The relaxation step shifts the focus from the strictly score-based view in \eqref{eq:argmaxMPL} towards a constraint- or regression-based view, or in terms of dependency networks, from symmetric to general. It is worth noticing that the consistency result established in Theorem \ref{thm:MPLconsistency} still holds under the relaxed problem formulation.

By solving the relaxed problem we usually obtain a solution inconsistent with a Markov network structure. We could simply post-process the solution using either a $\wedge$ (and) criterion,
\[
E_{\wedge}=\{ \{ i,j \} \in \{V \times V\} : i\in mb(j) \wedge j \in mb(i)   \}
\] 
or a $\vee$ (or) criterion,
\[
E_{\vee}=\{ \{ i,j \} \in \{V \times V\} : i\in mb(j) \vee j \in mb(i)   \}.
\]
These criteria are quite standard among constraint- or regression-based methods, however, neither of them is quite satisfactory from an MPL optimization perspective. Therefore we propose a second optimization phase whose goal is to combine the inconsistent Markov blankets from the first phase into a coherent structure which is MPL-optimal on a reduced model space determined by the relaxed solution.

More specifically, the edge set in $E_{\vee}$ is considered to be the result of a prescan that identifies eligible edges. The original problem \eqref{eq:argmaxMPL} is then solved with respect to the reduced model space $\mathcal{G}_{\vee}=\{ G\in\mathcal{G}:E\subseteq E_{\vee} \}$, that is
\begin{equation*}
\underset{G\in\mathcal{G}_{\vee}}{\arg \max} \ \log \hat p(\mathbf{x}\mid G).
\end{equation*}
The reduced model space $\mathcal{G}_{\vee}$ is in general considerably smaller than $\mathcal{G}$. In Section \ref{sec:PBO} we discuss a method that under certain circumstances can solve the above problem exactly. In Section \ref{sec:HC} we describe a fast deterministic approximate algorithm that can be applied also in situations when the exact method is infeasible.  

\subsection{Local Markov blanket discovery using greedy hill climbing\label{sec:MB_disc}}
To solve the relaxed problem, we basically need a Markov blanket discovery algorithm whose goal is to optimize the local MPL for each node independently of the solutions of the other nodes. For this we use an approximate deterministic hill climbing procedure similar to the interIAMB algorithm by \citet{Tsamardinos2003}.

\begin{algorithm}[tb]\small{
\textbf{Procedure} Markov-Blanket-Hill-Climb(

\hspace{1cm}$j$,\phantom{$\mathbf{x}$,} \ \ //\emph{Current node}

\hspace{1cm}$\mathbf{x}$,\phantom{$j$,} \ \ //\emph{Complete dataset}

\hspace{1cm})

1: \ \ $mb(j),\skew{6}\widehat{m}b(j)\leftarrow\varnothing$\vspace{0.1cm} 

2: \ \ \textbf{while} $\skew{6}\widehat{m}b(j)$ has changed \vspace{0.1cm}

3: \hspace{1cm} $C\leftarrow V\setminus \{ mb(j)\cup j\}$\vspace{0.1cm} 

4: \hspace{1cm} $mb(j)\leftarrow \skew{6}\widehat{m}b(j)$\vspace{0.1cm} 

5: \hspace{1cm} \textbf{for each} $i\in C$\vspace{0.1cm}  

6: \hspace{2cm} \textbf{if} $\log p(\mathbf{x}_{j}\mid \mathbf{x}_{mb(j)\cup i}) > \log p(\mathbf{x}_{j}\mid \mathbf{x}_{\skew{6}\widehat{m}b(j)})$\vspace{0.1cm} 

7: \hspace{3cm} $\skew{6}\widehat{m}b(j) \leftarrow mb(j)\cup i$\vspace{0.1cm}

8: \hspace{2cm} \textbf{end}\vspace{0.1cm} 

9: \hspace{1cm} \textbf{end}\vspace{0.1cm} 

10:\hspace{1cm} \textbf{while} $\skew{6}\widehat{m}b(j)$ has changed \& $|\skew{6}\widehat{m}b(j)|>2$\vspace{0.1cm} 

11:\hspace{2cm} $mb(j)\leftarrow \skew{6}\widehat{m}b(j)$\vspace{0.1cm} 

12:\hspace{2cm} \textbf{for each} $i\in mb(j)$\vspace{0.1cm}

13:\hspace{3cm} \textbf{if} $\log p(\mathbf{x}_{j}\mid \mathbf{x}_{mb(j)\setminus i}) > \log p(\mathbf{x}_{j}\mid \mathbf{x}_{\skew{6}\widehat{m}b(j)})$\vspace{0.1cm}

14:\hspace{4cm} $\skew{6}\widehat{m}b(j) \leftarrow mb(j)\setminus i$\vspace{0.1cm}

15:\hspace{3cm} \textbf{end}\vspace{0.1cm} 	

16:\hspace{2cm} \textbf{end}\vspace{0.1cm} 

17:\hspace{1cm} \textbf{end}\vspace{0.1cm} 

18:\ \ \textbf{end}\vspace{0.1cm} 

19:\ \ \textbf{return} $\skew{6}\widehat{m}b(j)$\vspace{0.1cm}} 
\caption{Procedure for optimizing the local MPL of a node using greedy hill climbing.\label{alg:findMB}}
\end{algorithm}

An outline of the algorithm is presented in Algorithm \ref{alg:findMB} and the general idea is as follows. The algorithm is based on the two basic operations by which members are added to or deleted from the Markov blanket. The method is initiated with the empty Markov blanket and all other nodes are considered potential Markov blanket members. At each iteration it adds to the Markov blanket the node that induces the greatest improvement to the local MPL and updates the set of potential members accordingly. When the size of the Markov blanket grows larger than two, the algorithm interleaves each successful addition-step with a deletion phase. In the deletion phase, the algorithm removes the node that induces the largest improvement to score. The deletion-step is repeated until removal of a node no longer increases the score or the size of the Markov blanket is smaller than three. When the addition-phase is iterated through without a successful addition, a local maximum has been reached, the algorithm terminates and returns the identified Markov blanket.

To examine the computational complexity of the algorithm, we consider the cost of performing an complete iteration where $mb$ denotes the current Markov blanket. In the addition phase, each of the $d-1-|mb|$ candidate members needs to be evaluated by calculating the local MPL for Markov blankets of size $|mb|+1$. Say that a node is added to the Markov blanket which is now of size $|mb|+1$. In the first iteration of a potential deletion phase, the removal of each of the $|mb|+1$ Markov blanket members is evaluated by calculating the local MPL for Markov blankets of size $|mb|$. In practice, the most lately added node can be skipped in the first iteration. In a potential successive deletion step, $|mb|$ Markov blankets of size $|mb|-1$ must be evaluated and so on.

The recurring deletion-phase of the algorithm attempts to keep the size of the Markov blanket as small as possible during the search in order to improve both sample and time efficiency. Still, the computational cost of the method is strongly dependent on the size of identified Markov blanket since a large Markov blanket naturally requires many iterations. Furthermore, an iteration becomes more expensive as the current Markov blanket grows larger since the cost of evaluating the local MPL is highly dependent on the size of the Markov blanket (see Section \ref{sec:MPLcomplex}). 

\subsection{Global graph discovery using pseudo-boolean optimization\label{sec:PBO}}
There has recently been a considerable interest in use of computational logic algorithms for structure learning of both Bayesian and Markov networks \citep{Cussens2008,BarlettCussens2013,CoranderJRNP2013,Jarvisalo2014,Parviainen2014}. In this section we describe how 
\begin{equation}\label{eq:PBO_original}
\underset{G\in\mathcal{G}_{\vee}}{\arg \max} \ \log \hat p(\mathbf{x}\mid G)
\end{equation}
can be cast as a pseudo-boolean optimization (PBO) problem \citep{Boros2002} which can be solved by existing mixed integer programming solvers such as the SCIP solver \citep{Berthold2009,Achterberg2009}.

A PBO problem consists of an objective function and a set of (in)equality constraints over boolean variables. To formulate our optimization problem as a PBO problem, we need to introduce two types of propositional variables: 
\begin{enumerate}
\item \textbf{Edge variables}: For each edge $\{i,j\} \in E_{\vee}$, a variable $x_{\{i,j\}}$ is introduced. If the value of $x_{\{i,j\}}$ is 1 (true) in a solution, the associated edge is included in the graph. If the value of $x_{\{i,j\}}$ is 0 (false), the associated edge is not included in the graph.

\item \textbf{Markov blanket variables}: Each node $j$ is associated with a set of candidate Markov blankets defined as all subsets of $mb_{\vee}(j)$ which is the Markov blanket of node $j$ in $G_{\vee}$.  Let $d_j$ be the number of nodes in $mb_{\vee}(j)$. The Markov blanket candidates are denoted by $mb_{k}(j)$ for $k=1,\ldots,m_{j}$ where $m_{j}=2^{d_{j}}$. For each candidate Markov blanket, a variable $x_{mb_{k}(j)}$ is introduced.  If the value of $x_{mb_{k}(j)}$ is 1 (true) in a solution, the Markov blanket of node $j$ is equal to $mb_{k}(j)$ in the graph. If the value of $x_{mb_{k}(j)}$ is 0 (false), the Markov blanket of node $j$ is not equal to the $k$:th candidate.  
\end{enumerate}
Each complete instantiation of the edge variables will correspond to a distinct graph in the considered graph space. The purpose of the edge variables is to ensure that the combined Markov blankets correspond to a coherent graph structure. Consequently, we need to connect the edge variables to the blanket variables in such a way that the value of blanket variable is true if and only if all edge variables associated with edges induced by the Markov blanket are true and the remaining edge variables are false. More formally, we need to introduce a constraint corresponding to the propositional formula
\begin{equation} \label{eq:PBO1}
x_{mb_{k}(j)} \leftrightarrow (x_{\{v_1,j\}} \wedge x_{\{v_2,j\}} \wedge \dots \wedge x_{\{v_l,j\}} \wedge \neg{x}_{\{v_{l+1},j\}} \wedge \neg{x}_{\{v_{l+2},j\}} \wedge \dots \wedge \neg{x}_{\{v_{d_j},j\}})
\end{equation}
where 
\begin{equation*}
\{v_1,\ldots,v_{l}\} = mb_{k}(j) \text{ and } \{v_{l+1},\ldots,v_{d_{j}}\} = mb_{\vee}(j)\setminus mb_{k}(j).
\end{equation*}
If we now consider the variables taking on values 0 and 1 rather than false and true, the above formula can be expressed as the pseudo-boolean equality constraint
\begin{equation} \label{eq:PBO2}
x_{mb_{k}(j)} - \left(\prod_{i=1}^{l} x_{\{v_i,j\}}\right) \left( \prod_{i=l+1}^{d_{j}} \bar{x}_{\{v_i,j\}}\right) = 0,
\end{equation}
where $\bar{x}_{\{v_i,j\}}=1-x_{\{v_i,j\}}$. For any assignment of the variables, it is clear that the value of formula \eqref{eq:PBO1} is true if and only if constraint \eqref{eq:PBO2} is satisfied. For each node and Markov blanket candidate, we add constraint \eqref{eq:PBO2} to the PBO problem. This will ensure that any feasible instantiation of the introduced variables must coincide with a graph structure of a Markov network.

The constraints expressed by equation \eqref{eq:PBO2} are sufficient on their own, however, to facilitate the optimization process we also introduce the following constraint for each node: 
\begin{equation} \label{eq:PBO3}
\sum_{k=1}^{m_{j}}  x_{mb_{k}(j)} = 1
\end{equation}
By including constraint \eqref{eq:PBO3}, we explicitly require that exactly one candidate is selected for each node. Even though this is already implied by constraints (16), the implication is not straightforward since the candidate variables are related to each other via the edge variables. Consequently, to realize that any two given candidate variables of the same node can not be true simultaneously, some of the edge variables need to be assigned. Therefore, including constraint (17) helps the solver to tighten the bounds of the feasible region (and objective function) early on.

Finally, we need to express our objective function. For this we introduce the Markov blanket candidate weights 
\begin{equation*}
w(j,k)=-\lfloor K \cdot \log p(\mathbf{x}_{j}|\mathbf{x}_{mb_{k}(j)})\rfloor
\end{equation*}
where $K$ is a large positive integer. As required in a PBO objective function, the floor function transforms the weights into integers. The objective function to be minimized can now be expressed by 
\begin{equation} \label{eq:PBO4}
\sum_{j=1}^d \sum_{k=1}^{m_{j}} w(j,k)\cdot x_{mb_{k}(j)}.
\end{equation}
With a large enough $K$, the solution to the PBO problem
\begin{equation} \label{eq:PBO5}
\underset{j=1,\ldots,d}{\underset{mb(j)\subseteq mb_{\vee}(j)}{\arg \min}} \ \sum_{j=1}^d \sum_{k=1}^{m_{j}} w(j,k)\cdot x_{mb_{k}(j)},
\end{equation}
subject to constraint \eqref{eq:PBO2} (and \eqref{eq:PBO3}), is equivalent to the solution to the optimization problem
\begin{equation} \label{eq:PBO5}
\underset{j=1,\ldots,d}{\underset{mb(j)\subseteq mb_{\vee}(j)}{\arg \min}} \ -K\sum_{j=1}^d \log p(\mathbf{x}_{j}\mid \mathbf{x}_{mb(j)}),
\end{equation}
subject to the constraint in \eqref{eq:argmaxMPLrf}, which in turn is equivalent to the original optimization problem \eqref{eq:PBO_original}.

The obvious advantage of this approach is that we are guaranteed to obtain the exact (or global) solution to the problem. However, the method can only be applied in certain situations since the number of variables and constraints for each node grows exponentially with the number of the potential Markov blanket members. More specifically, the total number of introduced boolean variables is $|E_{\vee}|+\sum_{j=1}^{d}2^{d_j}$ and the total number of introduced equality constraints is $d+\sum_{j=1}^{d}2^{d_j}$. In addition, the weight of each candidate must be calculated and stored prior to the actual optimization. Consequently, the feasibility of this approach depends strongly on the sizes of the Markov blankets in $G_{\vee}$. Hence, in the next section we also introduce an alternative approximate algorithm that can be applied also in intractable situations.

\subsection{Global graph discovery using greedy hill climbing\label{sec:HC}}
The variable-wise factorization of the MPL makes it particularly well-suited for global search algorithms based on local changes. As a stochastic option, the non-reversible MCMC-based approach by \citet{Corander2008} is directly applicable for MPL optimization. Here we propose a simple deterministic approach in form of a greedy hill climbing (HC) algorithm which has also been used for learning Bayesian networks \citep[see e.g.][]{Heckerman1995}. Local edge change algorithms move between neighboring graph structures during the optimization procedure. The set of neighbors of a graph $G$ in a graph space $\mathcal{G}$ is denoted by $\mathcal{N}_{\mathcal{G}}(G)$ and defined as all graphs in $\mathcal{G}$ that can be reached from $G$ by a adding or removing a single edge.
\begin{algorithm}\small{
\textbf{Procedure} Graph-Hill-Climb(

\hspace{1cm}$\mathcal{G}_{\vee}$\phantom{$\mathbf{x}$,} \ \ //\emph{The considered graph space}

\hspace{1cm}$\mathbf{x}$,\phantom{$\mathcal{G}_{\vee}$} \ \ //\emph{Complete dataset}

\hspace{1cm})

1: \ \ $G,\widehat{G}\leftarrow\varnothing$\vspace{0.1cm} 

2: \ \ \textbf{while} $\widehat{G}$ has changed\vspace{0.1cm}

3: \hspace{1cm} $G\leftarrow \widehat{G}$\vspace{0.1cm} 

4: \hspace{1cm} \textbf{for each} $G'\in \mathcal{N}_{\mathcal{G}_{\vee}}(G)$\vspace{0.1cm}  

5: \hspace{2cm} \textbf{if} $\hat p(\mathbf{x}\mid G') > \hat p(\mathbf{x} \mid \widehat{G})$\vspace{0.1cm} 

6: \hspace{3cm} $\widehat{G} \leftarrow G'$\vspace{0.1cm} 

7: \hspace{2cm} \textbf{end}\vspace{0.1cm} 

8: \hspace{1cm} \textbf{end}\vspace{0.1cm} 

9: \ \ \textbf{end}\vspace{0.1cm} 

10:\ \ \textbf{return} $\widehat{G}$\vspace{0.1cm}} 
\caption{Procedure for optimizing the MPL using greedy hill climbing.\label{alg:HC}}
\end{algorithm}

An outline of the algorithm is presented in Algorithm \ref{alg:HC} and the general idea is as follows. The empty graph is set as the initial graph and the considered optimization space is $\mathcal{G}_{\vee}$. At each iteration, all neighbors of the current graph are evaluated. At the end of the iteration, we choose the highest scoring graph from the neighbors, assuming that it has a higher score than the current graph, and repeat the procedure. If no candidate among the neighbors has a higher score than the current graph, a local maximum has been reached, the algorithm terminates and returns the identified graph.

We examine the computational complexity of the proposed algorithm by considering the calculations required at each iteration. The cost of evaluating the specified expressions was discussed in Section \ref{sec:MPLcomplex}. We show that, by implementing smart caching, the efficiency of the algorithm can be improved considerably. Let $G_{t}$ be the current graph at iteration $t$ and let $G'_{t}\in \mathcal{N}_{\mathcal{G}_{\vee}}(G_{t})$ differ with respect to the edge $\{i,j\}$. To compare $G'_{t}$ to $G_{t}$ we calculate the log-Bayes pseudo-factor   
\begin{equation*}
\begin{aligned}
\log K(G^{*}_{t},G_{t})=	&\log p(\mathbf{x}_{i} \mid \mathbf{x}_{mb(i)},G'_{t})+\log p(\mathbf{x}_{j} \mid \mathbf{x}_{mb(j)},G'_{t})-\\
				&\log p(\mathbf{x}_{i} \mid \mathbf{x}_{mb(i)},G_{t})-\log p(\mathbf{x}_{j} \mid \mathbf{x}_{mb(j)},G_{t}). 
\end{aligned}
\end{equation*}
The above expression must be evaluated for each candidate in the neighbor set $\mathcal{N}_{\mathcal{G}_{\vee}}(G_{t})$ which has a maximum cardinality of $d \choose 2$ if all edges are included. However, since 
\[
\log p(\mathbf{x}_{i} \mid \mathbf{x}_{mb(i)},G_{t}) \text{ and } \log p(\mathbf{x}_{j} \mid \mathbf{x}_{mb(j)},G_{t}) 
\]
are determined by the current graph, they can be stored and re-used. Consequently,
 \[
\log p(\mathbf{x}_{i} \mid \mathbf{x}_{mb(i)},G'_{t}) \text{ and } \log p(\mathbf{x}_{j} \mid \mathbf{x}_{mb(j)},G'_{t}) 
\]
are the only terms that specifically need to be calculated to evaluate the neighbor associated with the edge change $\{i,j\}$. 

In the first iteration, $d$ local MPLs with empty Markov blankets must be evaluated. Additionally, each possible neighbor must be evaluated by calculating two local MPLs with Markov blanket size one.  In subsequent iterations, we can further exploit the decomposition of the MPL by noting that most of the log-factors from the previous iteration remain unchanged under the new current graph. In fact, the only edge changes that need to be re-evaluated are those that overlap with the previous change. Say that the current graph $G_{t}$ was attained by adding (deleting) the edge between $\{k,l\}$ to (from) $G_{t-1}$. In line with earlier notation, let $G'_{t-1}$ be the neighbor of $G_{t-1}$ that differs with respect to the edge $\{i,j\}$. Given the context, we now have that
\[
\log K(G'_{t},G_{t})=\log K(G'_{t-1},G_{t-1}) \text{ if } \{i,j \}\cap \{k,l \}=\varnothing.
\]
Consequently, after the initial iteration it suffices to re-evaluate only a small fraction of the updated neighbor set. In the worst case, $2(d-1)$ neighbors need to be evaluated even though the maximum cardinality of the neighbor set of interest is ${d \choose 2}-1$ when excluding the graph from the previous iteration. 

Under this optimization strategy, the MPL method is similar in spirit to the max-min hill climbing algorithm for learning Bayesian networks by \citet{Tsamardinos2006}. The main difference is that both phases of our algorithm are derived from the notion of maximizing a single underlying score.

\section{Experimental results}\label{sec:Results}
The main focus of this section is to empirically investigate the performance of the MPL using the optimization algorithms from the previous section. To evaluate our approach in a controlled setting, we compare it to other competitive methods on synthetic models as well as real-world networks. Since the graph structures of the generating models are known, it allows for a straightforward and fair assessment of the algorithms. To finally illustrate the potential of our method, we also present a real high-dimensional knowledge discovery problem on which our method is applied.

When the true structure of the generating network is known, the quality of an output network is readily assessed by the number of errors in terms of FPs and FNs. As our main measure of quality we consider the sum of FPs and FNs which is the Hamming distance between the output and true network. Consequently, a low value on the Hamming distance corresponds to structural resemblance to the true network and the minimum value of zero is obtained for the correct graph. In addition to structural resemblance, we monitor the execution times for the different methods.\footnote{All experiments were carried out in Matlab except for the PBO, which was solved using the SCIP solver (web site: {\tt{http://scip.zib.de/}}). The experiments were performed on a standard PC architecture with 2.66GHz dual-core processors.} The total runtimes of all algorithms are reported along with the maximum discovery time for a single Markov blanket. The maximum Markov blanket discovery time would be the total real time required if the local problems were solved in parallel rather than in serial fashion.

If not otherwise mentioned, we set the equivalent sample size parameter $N=1$ which results in a weak parameter prior. For the main part of the experiments, we set $p(G)$ to be uniform since we want to investigate how well the MPL alone performs as a metric for graph structures.

\subsection{Synthetic Markov networks\label{sec:SYNTH_SIM}}
In this section we use synthetic models to generate datasets of different sizes to systematically compare the performance of the MPL combined with our optimization algorithms. Moreover, we compare the MPL against different competing methods for structure learning of Markov networks. For simplicity, we restrict the synthetic networks to be made up of binary variables.
\begin{figure}[tb]
\begin{subfigure}{0.24\textwidth}
\begin{center}
\includegraphics[width=3cm]{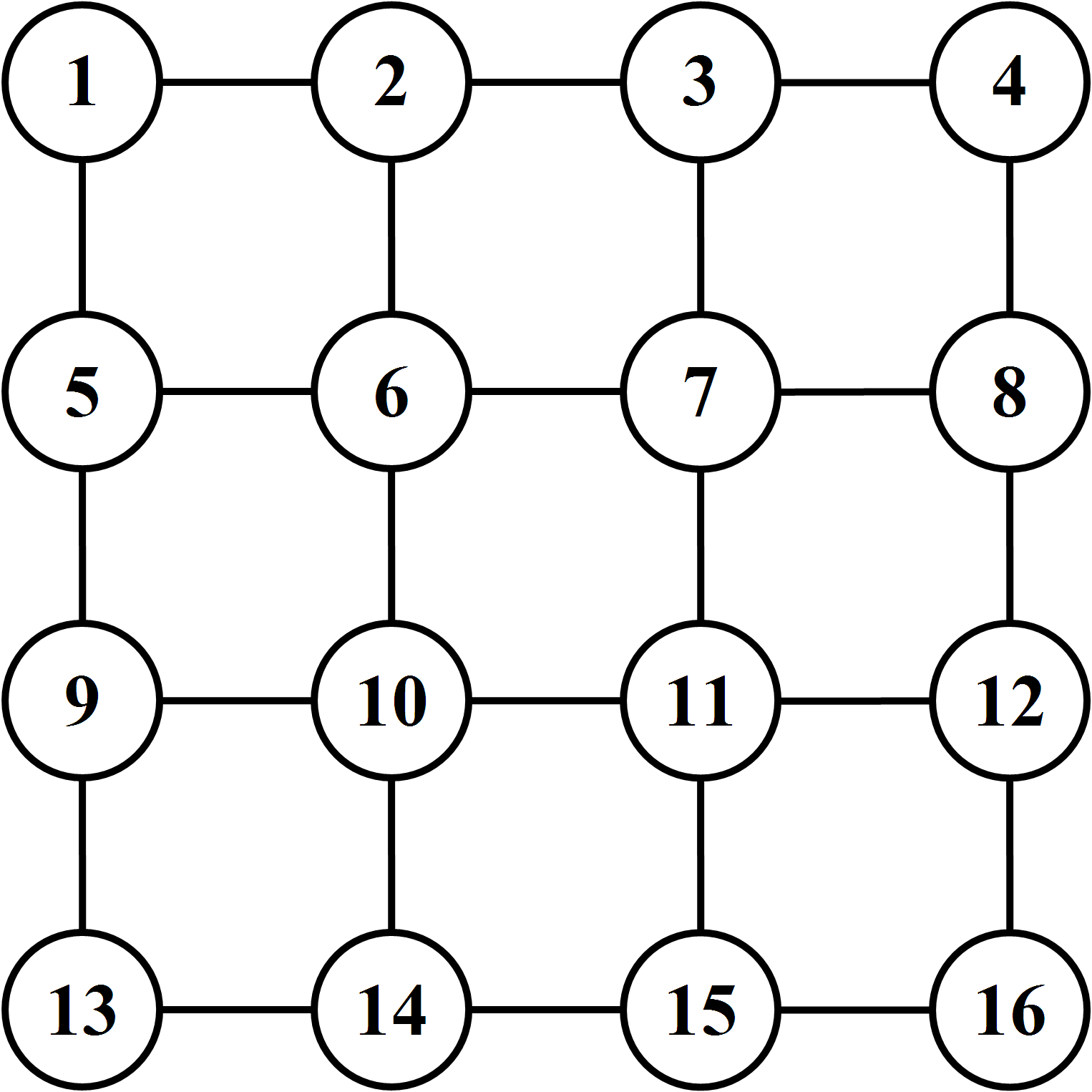}
\caption{}
\end{center}
\end{subfigure}
\begin{subfigure}{0.24\textwidth}
\begin{center}
\includegraphics[width=3cm]{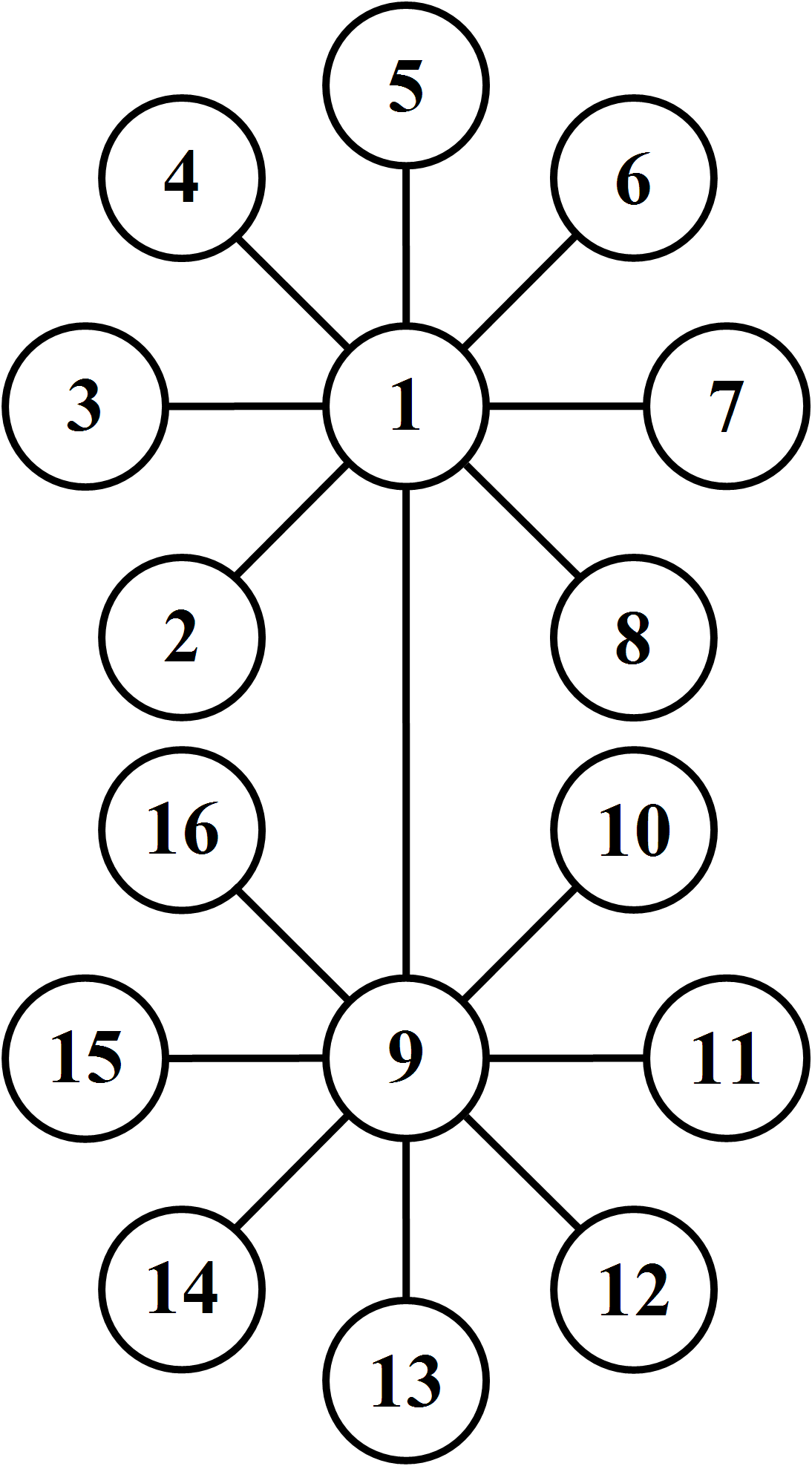}
\caption{}
\label{fig:d16hub}
\end{center}
\end{subfigure}
\begin{subfigure}{0.24\textwidth}
\begin{center}
\includegraphics[width=3cm]{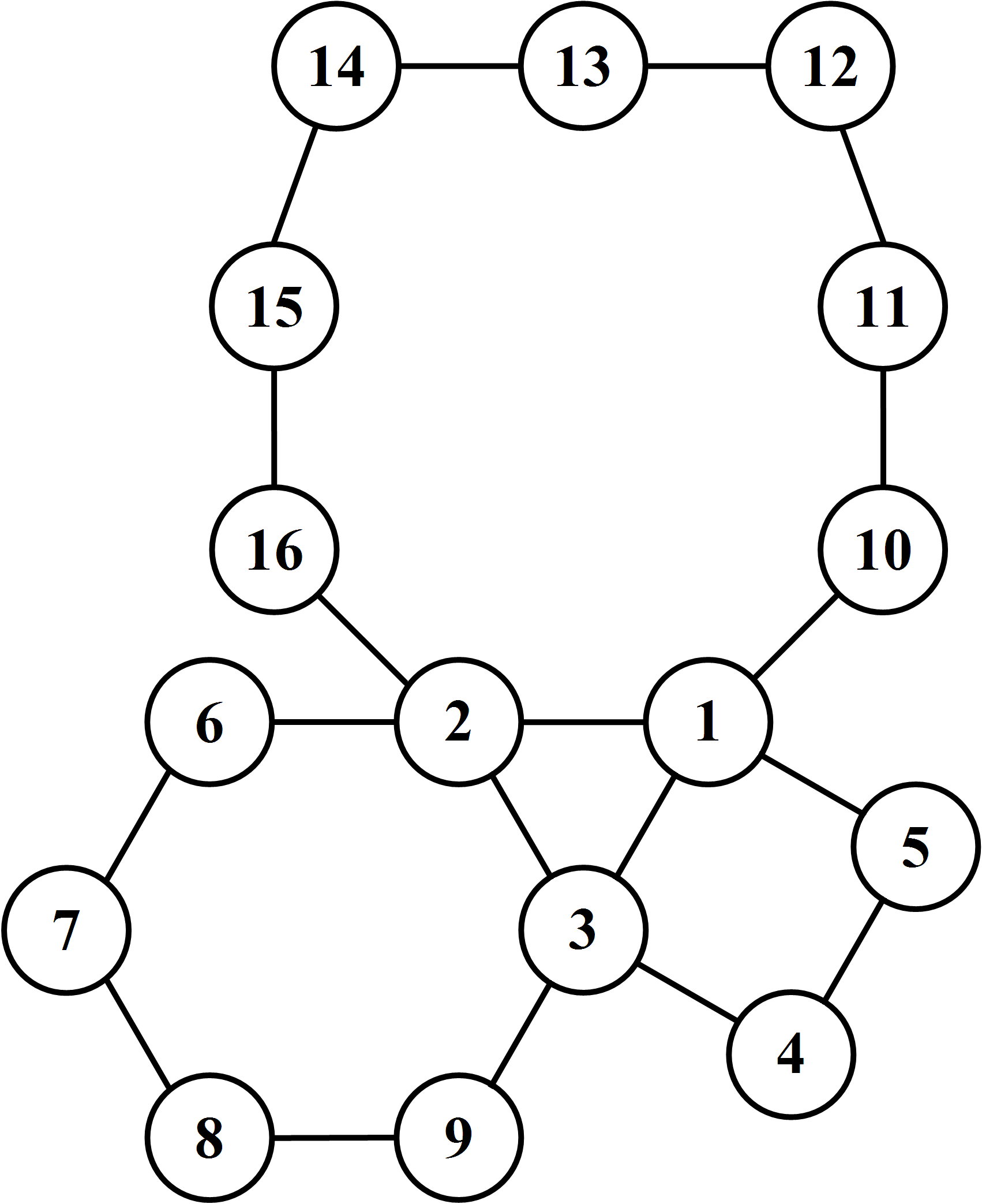}
\caption{}
\end{center}
\end{subfigure}
\begin{subfigure}{0.24\textwidth}
\begin{center}
\includegraphics[width=3cm]{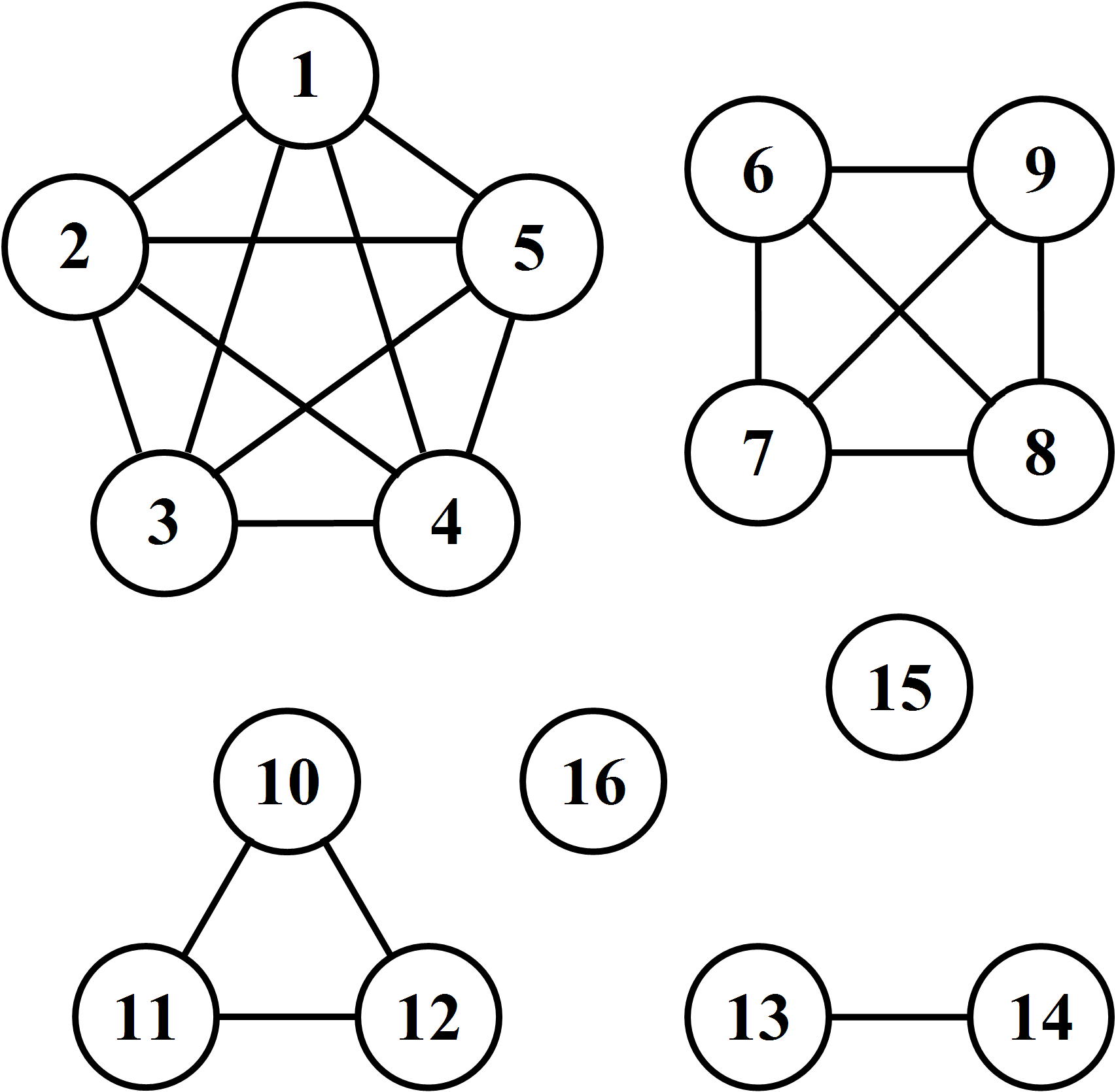}
\caption{}
\end{center}
\end{subfigure}
\caption{Synthetic graphs used in Section \ref{sec:SYNTH_SIM}: (a) Grid, (b) Hub, (c) Loop, and (d) Clique.}
\label{fig:d16comp}
\end{figure}
\begin{table}
\captionsetup{skip=-5pt}
\small{
\begin{center}
\begin{tabular}{l | c c c c }
Network & Grid & Hub & Loop & Clique \\
\hline
Number of nodes& 16 & 16 & 16 & 16 \\
Number of edges& 24 & 15 & 19 & 20 \\
Average Markov blanket size & 3.25 & 1.88 & 2.38 & 2.5 \\
Maximum Markov blanket size & 4 & 8 & 4 & 4 \\
Chordal & No & Yes & No & Yes \\
\end{tabular}
\end{center}
}
\caption{Properties of the graph components in Figure \ref{fig:d16comp}. \label{tab:d16comp}}
\end{table}

The synthetic graphs were formed by combining disconnected components in form of the four 16-node graphs illustrated in Figure \ref{fig:d16comp}. These graphs represent different structural characteristics present in realistic models and some of their properties are listed in Table \ref{tab:d16comp}. In particular, as already shown in Section \ref{sec:Properties}, the hub network in Figure \ref{fig:d16hub} represents a structural characteristic that is especially hard to capture for the MPL even though it is a rather simple tree structure. Initially, one replica of each subgraph was combined to form a structure over 64 variables. This procedure was then repeated with 2, 4, and 8 replicas to form network structures over 128, 256, and 512 variables, respectively. Each final network structure thus contained all the structural characteristics present in the graph components. The advantage of this approach is that the disconnected nature of the generating networks facilitates the sampling procedure substantially since each distinct subnetwork can be sampled directly from its corresponding joint distribution independently of the rest of the network. In practice, a distribution was generated by randomly sampling the maximal clique factors in \eqref{eq:cliqueFAC}. Each factor value $\phi(x_{C})$ was drawn, independently of the other values, from a uniform distribution over $(0,1)$. Consequently, the strength of the dependencies entailed by the edges may have varied considerably. To increase the stability of our results, for each sample size and graph structure, we generated 10 distributions from each of which 10 datasets were sampled. In total, under each setup, we learned 100 model structures over which the final results were averaged. The experiments were performed for sample sizes ranging from $250$ to $32000$.

First we examine how our approximate algorithm, HC, performs in comparison with our exact algorithm, PBO, in the second phase of the optimization process. Since the exact method finds the globally MPL-optimal graph on the reduced graph space $\mathcal{G}_{\vee}$, we can use it as a gold standard to which we compare our approximate method. Since the feasibility of the exact method is restricted by the output of the first phase, we need to filter out instances that are not solvable in a reasonable time. We restricted the comparison to instances where the total number of Markov blanket candidates is less than 15000. We also set a time limit on the solver to 3600 seconds per instance. The following results are thus based on the remaining solved instances (see Table \ref{tab:PBO_HC_COMP} for more details). 
\begin{figure}
\begin{subfigure}{0.5\textwidth}
\captionsetup{skip=-5pt}
\begin{center}
\includegraphics{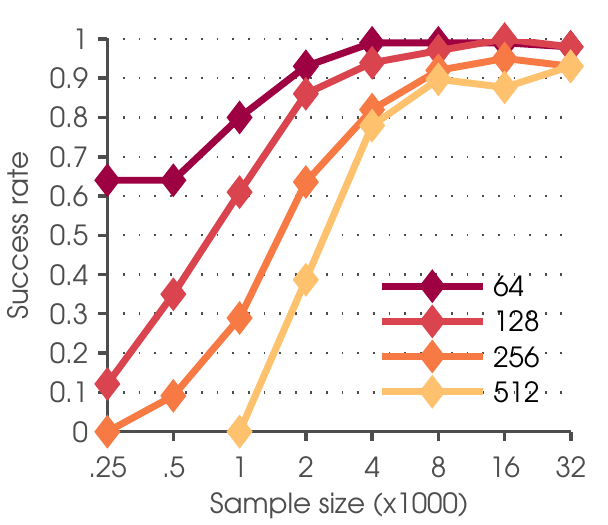}
\end{center}
\caption{\label{fig:PBOvsHCa}}
\end{subfigure}
\begin{subfigure}{0.5\textwidth}
\captionsetup{skip=-5pt}
\begin{center}
\includegraphics{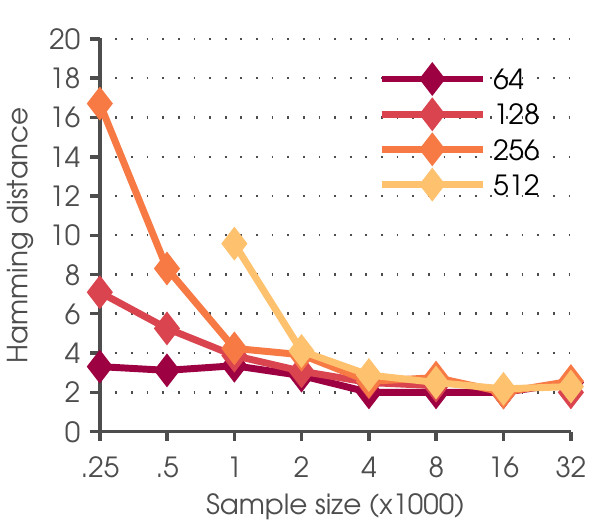}
\end{center}
\caption{\label{fig:PBOvsHCb}}
\end{subfigure}
\caption{Comparison of HC and PBO for different model sizes. In (a) the rate at which the two methods reach identical solutions is plotted against the sample size. In (b) the average Hamming distance between non-identical solutions is plotted against the sample size.}
\end{figure} 

In Figure \ref{fig:PBOvsHCa} we have plotted the rate at which the two methods discover identical solutions. In terms of the HC method, this corresponds to the rate at which the algorithm succeeds in reaching the global optimum. As expected, the success rate grows with an increased sample size. When given more observations the algorithm is more firmly guided towards the optimal solution. In addition, the size of the optimization space is in general smaller for larger sample sizes due to an increased conformity among the Markov blankets from the initial phase. Overall, the HC method performs very well in identifying the optimal solution for reasonable sample sizes. We now consider the instances where the optimal solution was not reached by the HC method. In Figure \ref{fig:PBOvsHCb} we have plotted the average Hamming distance between the HC solution and the optimal solution for instances when the two graphs were different. For all of the considered model sizes, the curves quickly converge towards a Hamming distance of two which is the closest a local maximum can be to the global maximum under our definition of neighboring graphs. As expected, the approximate and exact solutions tend to resemble each other somewhat less for the more extreme ``large $d$, small $n$''-setups. 

In the second part of the experiment, we compare the MPL against other structure learning methods that are also applicable in high dimensions. We limit the MPL approach to the less computationally expensive HC algorithm, which we from now on simply refer to as the MPL method. The other methods used in the comparison are the following:

\begin{itemize}
\item \textbf{PIC}: The PIC criterion by \citet{Csiszar2006} is applied using the exact same search technique as for the MPL method. From the proof of Theorem \eqref{thm:MPLconsistency}, we know that MPL and PIC are asymptotically equivalent estimators, but here we examine how they perform in practice for limited sample sizes.

\item \textbf{CMI}: We apply the Markov blanket discovery approach of \cite{Tsamardinos2003} who use conditional mutual information\footnote{To calculate the conditional mutual information, the Matlab package of \cite{HanchuanPengMIpackage} was used.} (CMI) to assess if two variables are conditionally independent given some set of variables. For a fair comparison against the MPL method, we use the CMI measure combined with Algorithm \ref{alg:findMB}. To form the final graph, we apply either the AND or the OR criterion depending on which one results in a graph closer to the true graph in terms of Hamming distance.

\item \textbf{L1LR}: We apply the $L_{1}$-regularized logistic regression\footnote{The $L_{1}$-regularized logistic regression was performed using the Matlab package of \cite{MarkSchmidtMatlabCode}.} (L1LR) approach of \cite{Ravikumar2010} which is directly applicable on our models since we have restricted the experiments to binary variables. To form the final graph, we apply either the AND or the OR criterion depending on which one results in a graph closer to the true graph in terms of Hamming distance.
\end{itemize}
\begin{figure}[tb]
\begin{subfigure}{0.5\textwidth}
\captionsetup{skip=-5pt}
\begin{center}
\includegraphics{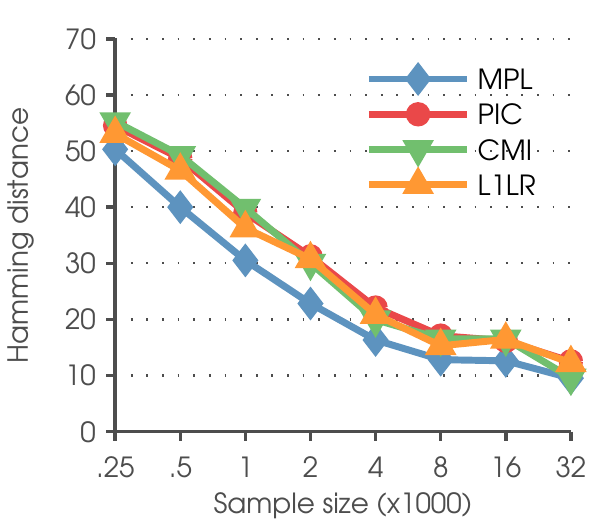}
\end{center}
\caption{}
\end{subfigure}
\begin{subfigure}{0.5\textwidth}
\captionsetup{skip=-5pt}
\begin{center}
\includegraphics{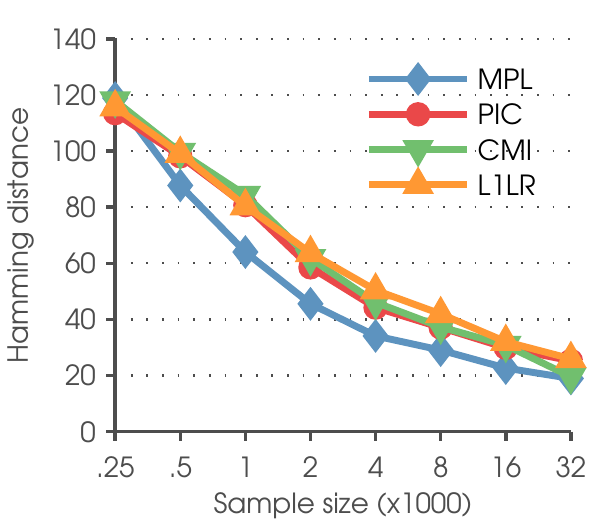}
\end{center}
\caption{}
\end{subfigure}
\begin{subfigure}{0.5\textwidth}
\captionsetup{skip=-5pt}
\begin{center}
\includegraphics{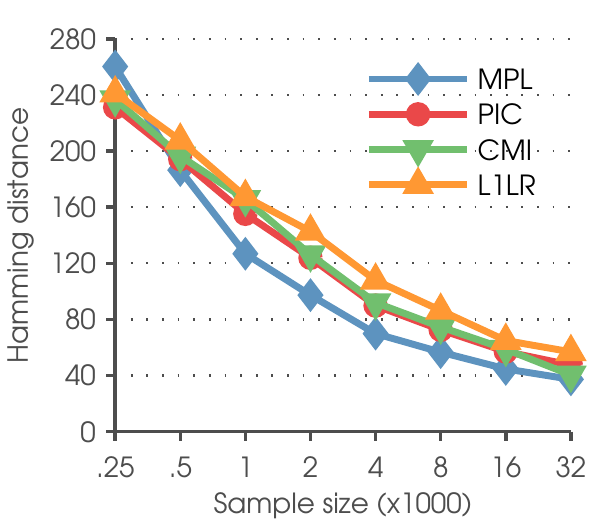}
\end{center}
\caption{}
\end{subfigure}
\begin{subfigure}{0.5\textwidth}
\captionsetup{skip=-5pt}
\begin{center}
\includegraphics{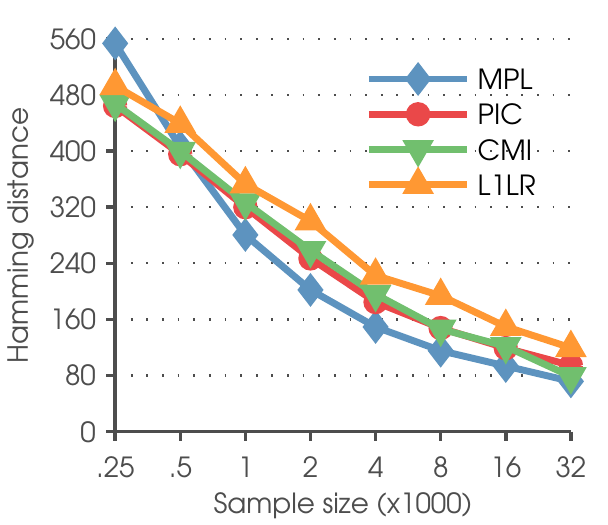}
\end{center}
\caption{}
\end{subfigure}
\caption{Comparison of the MPL method with the other structure learning methods. The average Hamming distance between the induced and true graph is plotted against the sample size for synthetic models of size (a) $d=64$, (b) $d=128$, (c) $d=256$, and (d) $d=512$.\label{fig:MPLvsOTHER}}
\end{figure} 
An issue with both the CMI and L1LR method is that they require the user to specify a crucial tuning parameter in form of a threshold value and a regularization weight, respectively. Consequently, to circumvent this problem, the methods were executed for the following ranges of values:
\[
\lambda_{\text{CMI}} \in \{0.25,0.1,0.075,0.05,0.025,0.01,0.0075,0.005,0.0025,0.001, 0.00075,0.0005\},
\]
\[
\lambda_{\text{L1LR}} \in \{4,6,8,12,16,24,32,48,64,96,128,192,256,384,512,768,1024\}.
\]
This resulted in a range of overly sparse to overly dense graphs from which we picked the graph (and parameter value), that minimized the Hamming distance, as the final solution. Since the Hamming distance tend to steadily increase when moving away from the optimal parameter value, the above range of values were chosen such that the picked parameter values (see Table \ref{tab:TH_REG}) would lie strictly between the smallest and largest value. In a sense, the MPL also requires us to choose a value on the equivalent sample size parameter $N$. However, in these experiments we have fixed $N=1$ whereas the other methods are tuned with respect to the true graph in order to perform optimally given the range of parameter values. 

The results of the simulations are summarized in Table \ref{tab:COMP_TPFP} and \ref{tab:COMP_TIMES}. In Figure \ref{fig:MPLvsOTHER} the average Hamming distance is illustrated for the different methods and model sizes. Overall, the MPL method performed highly satisfactorily and was marginally inferior to the other methods only for some of the most extreme ``large $d$, small $n$''-settings. It is difficult to say whether this is due to the MPL itself or the approximation made by the HC method. In terms of speed (see Table \ref{tab:COMP_TIMES}), the MPL method was at a comparable level for all of the considered models. Furthermore, the task of determining the tuning parameter experimentally would significantly increase the total runtimes of the CMI and L1LR method.

\subsection{Real-world Bayesian networks \label{sec:BM_SIM}}
\begin{table}[tb]
\small{
\begin{center}
\begin{tabular}{l | c c c c }
Network & Alarm & Insurance & Hailfinder & Barley \\
\hline
Number of nodes& 37 & 27 & 56 & 48 \\
Number of edges (DAG)& 46 & 52 & 66 & 84 \\
Number of edges (moral graph)  & 65 & 70 & 99 & 126 \\
Number of parameters & 509 & 984 & 2656 & 114005 \\
Average Markov blanket size & 3.51 & 5.19 & 3.54 & 5.25 \\
Maximum Markov blanket size & 8 & 10 & 17 & 13 \\
Average variable cardinality & 2.84 & 3.30 & 3.98 & 8.77 \\
Maximum variable cardinality & 4 & 5 & 11 & 67 \\
\end{tabular}
\end{center}
}
\caption{Properties of the real-world Bayesian networks used in Section \ref{sec:BM_SIM}.\label{tab:BMnetworks}}
\end{table}
In this section we proceed to a more realistic setting by conducting experiments on well-known real-world models, from the related class of Bayesian networks, in a similar fashion as \citet{Bromberg2009}. The considered models are commonly used as benchmarks in research and are available from a number of sources\footnote{The networks used in this work were obtained from the Bayesian network repository at {\tt{http://www.bnlearn.com/bnrepository/}} (Accessed 2014-08-07) and sampled using the R package of \citet{Scutari2010}.}. To transform the directed acyclic graph of a Bayesian network into a corresponding undirected graph of a Markov network, a two-step procedure known as moralization is used \citep[see][]{Lauritzen1996,Koller2009}. In the first step all parents of a common child are connected by an undirected edge if not already connected. In the second step the graph is made undirected by removing the direction of all directed edges. Although the local Markov property remains valid in the transformed network, some conditional independencies are lost in the moralization process due to the added edges. Consequently, the associated distribution is no longer faithful to the undirected graph making the graph identification more challenging.  

We selected the four medium-sized networks which are listed along with some of their properties in Table \ref{tab:BMnetworks}. Compared to the relatively simple and balanced synthetic networks in Section \ref{sec:SYNTH_SIM}, these models are more challenging due to their higher edge density and larger Markov blankets. In addition, large variable cardinalities also tend to have a negative effect on the learning time for methods such as the MPL. As before, we sampled each network for sample sizes ranging from $250$ to $32000$. For each network and sample size, we generated 100 samples over which the final results were averaged. We applied the same methods as in the previous section except for L1LR which without modifications is restricted to binary variables. In order to have a sufficient range of threshold values for the CMI method we added
\[
\{ 0.5,0.75,1,1.25,1.5,1.75,2,2.25,2.5 \}
\]
to the range of values from the previous section. 

\begin{figure}[tb]
\begin{subfigure}{0.5\textwidth}
\captionsetup{skip=-5pt}
\begin{center}
\includegraphics{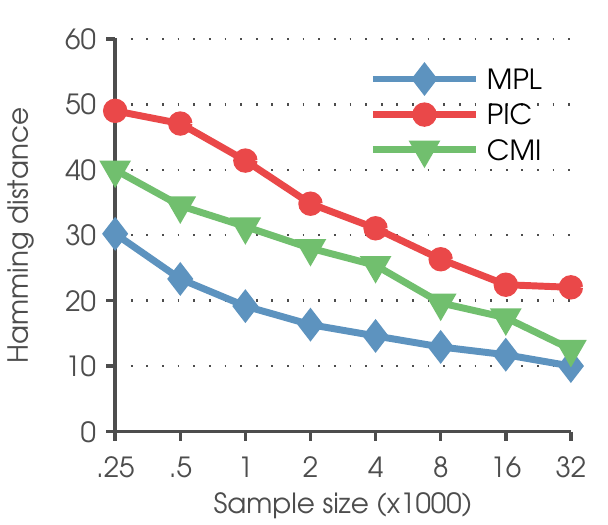}
\end{center}
\caption{}
\end{subfigure}
\begin{subfigure}{0.5\textwidth}
\captionsetup{skip=-5pt}
\begin{center}
\includegraphics{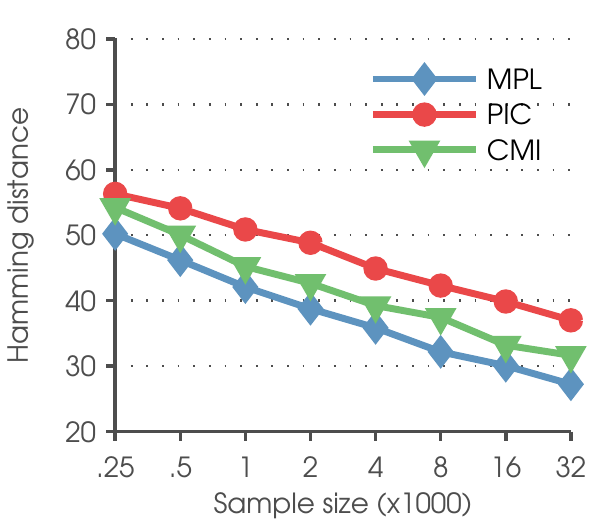}
\end{center}
\caption{}
\end{subfigure}
\begin{subfigure}{0.5\textwidth}
\captionsetup{skip=-5pt}
\begin{center}
\includegraphics{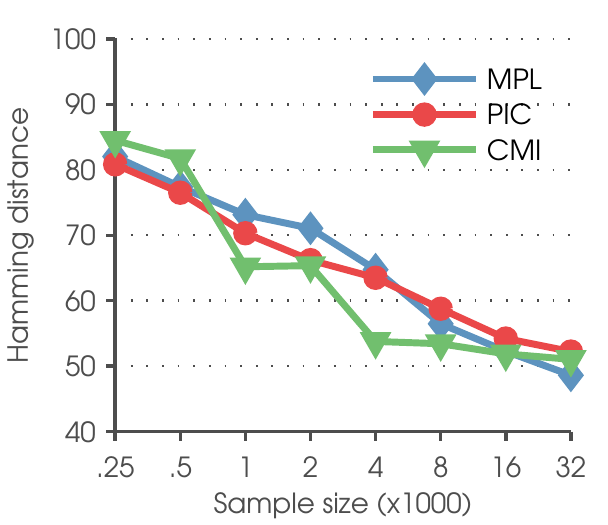}
\end{center}
\caption{}
\end{subfigure}
\begin{subfigure}{0.5\textwidth}
\captionsetup{skip=-5pt}
\begin{center}
\includegraphics{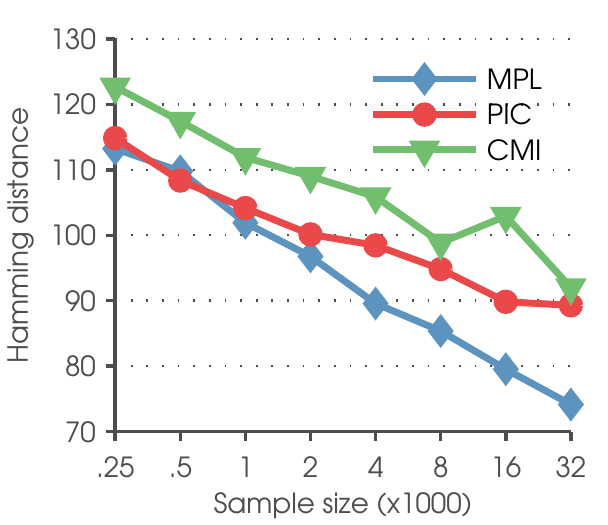}
\end{center}
\caption{}
\end{subfigure}
\caption{Comparison of the MPL method with the other structure learning methods. The average Hamming distance between the identified and true moral graph is plotted against the sample size for the (a) Alarm, (b) Insurance, (c) Hailfinder, and (d) Barley network.\label{fig:MPLvsOTHER_BM}}
\end{figure} 

The results of the simulations are summarized in Table \ref{tab:BM_TPFP} and \ref{tab:BM_TIMES}. In Figure \ref{fig:MPLvsOTHER_BM} the average Hamming distance from the moralized graph is illustrated for the different methods and networks. Again, the MPL method displayed an overall stable and good performance when compared to the other methods. However, the Hailfinder network exposes the main weakness of the MPL (and the related PIC when consistency is enforced). Although the majority of the nodes in the moral graph have relatively small Markov blankets, there is one node with a Markov blanket of size 17. As shown earlier, the MPL struggles with large Markov blankets or so-called hub nodes. As seen in Table \ref{tab:BM_TPFP}, the CMI-AND method naturally suffers from the same problem, however, the CMI-OR method can to some extent circumvent it. In terms of speed (see Table \ref{tab:BM_TIMES}), the MPL is the slowest of the considered methods, however, its runtimes are still at a reasonable level considering the performance. A reason for the slower runtimes is that the MPL method tends to produce denser graphs than the PIC as well as the CMI method when tuned with respect to the Hamming distance. Consequently, the MPL method requires more iterations which will affect the runtimes negatively, especially when the variable cardinalities are increased. However, it should be noted that the choice of threshold value by cross-validation would again significantly increase the computation time for CMI method.

All MPL simulations this far have been performed under the fixed equivalent sample size $N=1$. Whereas $\lambda_{\text{CMI}}$ and $\lambda_{\text{L1LR}}$ have a rather clear interpretation in terms of their effect on the graph, the effect of $N$ is not as easy to interpret. \citet{Silander2007} showed experimentally that the maximum a posteriori (MAP) structure of the BDeu score for Bayesian networks is sensitive to the choice of $N$. Furthermore, they noted that larger values of $N$ tend to produce denser MAP graphs. Since the MPL and BDeu share the same basic structure, one would expect to see a similar behavior between the two scores. To investigate this we conclude this section by performing an additional simulation study for the Alarm network for
\[
N \in \{ 1,4,8,16,32,64,128,256 \}.
\]
\begin{figure}[tb]
\captionsetup{skip=-5pt}
\begin{center}
\includegraphics{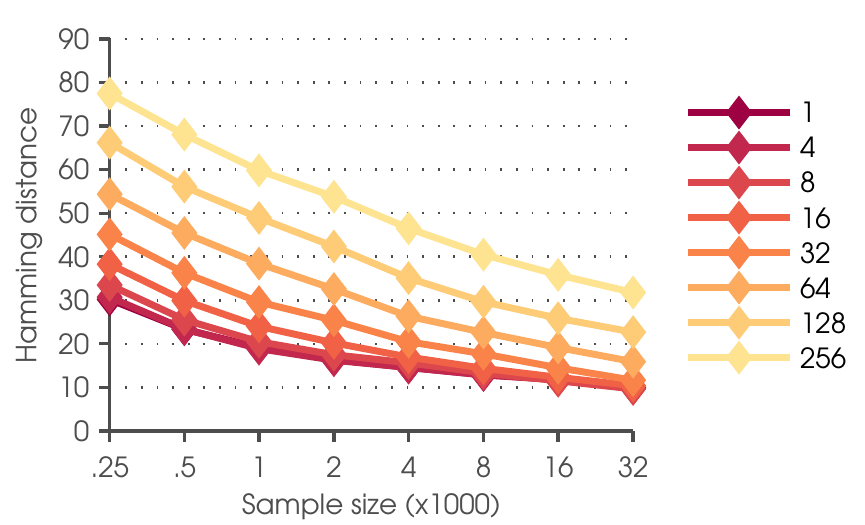}
\end{center}
\caption{The MPL method applied on the Alarm network under different values of $N$. The average Hamming distance between the identified and true moral graph is plotted against the sample size.\label{fig:ESS_HD}}
\end{figure} 

The results of the simulation are summarized in Table \ref{tab:ESS_TPFP} and the Hamming distances are illustrated in Figure \ref{fig:ESS_HD}. As expected, the results indicate a similar behavior as the BDeu metric in the sense that the MPL method produced denser graphs for larger values of $N$. Furthermore, the results in  Table \ref{tab:ESS_TPFP} also indicate that larger values of $N$ can be beneficial for larger samples, however, $N=1$ appears to be a reasonable choice when considering the complete range of sample sizes.

\subsection{A real-world application}\label{sec:GenEX}
Finally, to illustrate the MPL for a high-dimensional real application, we consider a dataset of 1,000 aligned whole-genome DNA sequences of \textit{Mycobacterium tuberculosis} \citep{Casali2014}. A Markov network can be used to reveal direct associations between variation over genome positions that may be relatively distant from each other. This purpose is similar to the use of Markov networks for finding direct dependencies among amino acid sequence positions that relate to the underlying crystal structure \citep{Aurell2013}. 
\begin{figure}[t]
\begin{center}
\includegraphics{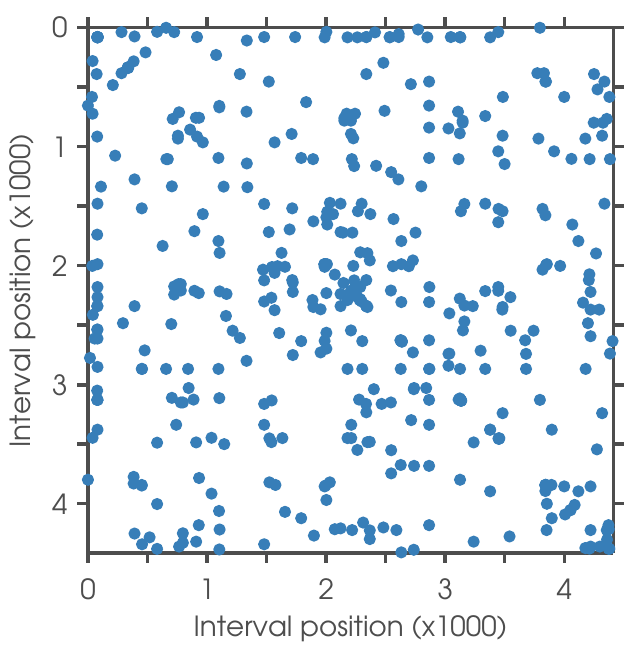}
\caption{Association pattern identified by the MPL method for 1kb genome intervals in \textit{M. tuberculosis}. A blue point indicates found linkage between positions that reside within the intervals located in the genome according to the values on the horizontal and vertical axes (in thousands of nucleotides).  }
\label{fig:linkDis}
\end{center}
\end{figure}

The original 5Mb multiple sequence alignment for \textit{M. tuberculosis} had approximately 27,000 variable positions, out of which we chose 589 that displayed sufficient variability determined by the threshold that the most frequent DNA base in a variable position was not allowed to represent more than $90\%$ of the total number of observations. Similar to the amino acid dependency modeling with Markov networks, associations between genome positions that are close neighbors in the sequences are trivial and uninteresting for the biological purposes. We applied the MPL method on the 589 variables under the sparsity promoting prior
\begin{equation*}
p(G)\sim \prod_{j=1}^{d} 2^{-q_{j}(r_{j}-1)}.
\end{equation*} 
in order to filter out the strongest dependencies.

The variable positions included in the analysis were separated maximally by almost 2.5 million bases since the bacterial genome is circular. To enable an efficient illustration of the results, the genome alignment was first split into non-overlapping intervals of 1000 (1kb) successive positions. Then, an adjacency matrix was created for the pairs of genome intervals such that a pair of intervals was determined adjacent if the identified graph contained an edge between any variable positions in the two intervals, respectively. The resulting adjacency structure is illustrated in Figure \ref{fig:linkDis}, which succinctly demonstrates the long-distance linkage of genome variation in this bacterium, as expected on the basis of its very low recombination rate. 

\section{Conclusions}
In this work we have introduced a novel approach for learning the structure of a Markov network without imposing the restriction of chordality. Our marginal pseudo-likelihood scoring method is proven to be consistent and can be considered a small sample analytical version of the information theoretic PIC criterion \citep{Csiszar2006}. We have designed two algorithms for finding the MPL-optimal solution after an initial prescan for plausible edges. For moderately sized candidate sets of Markov blankets, we have shown that it is possible to obtain an exact solution to the restricted global optimization problem using pseudo-boolean optimization. As a fast alternative to the exact method, we considered a greedy hill climbing approach, which gave near optimal performance for reasonable sample sizes. The straightforward possibility of parallel use of the MPL makes it a viable candidate for high-dimensional knowledge discovery.

In comparison with the other methods for structure learning of Markov networks, our MPL method was overall superior, and only slightly inferior to alternatives under fairly extreme ``large $d$, small $n$''-settings, or when the underlying network contained hub nodes. Moreover, it should also be kept in mind that we chose the tuning parameter values for the CMI and L1LR method by optimizing their performance against the known underlying structures, to reduce the computational burden of the experiments. In a real data analysis scenario, it would be necessary to tune these methods, using for example cross-validation, which would plausibly have a negative effect on their performance. In this sense, the comparison was extremely fair for the alternative methods, since no parameter choice was made for the MPL by the resulting performance. In terms of execution time, the MPL method is not necessarily as fast as the other methods, however, the runtimes of the CMI and L1LR method are here reported under a best case scenario. If the value of the penalty parameter would be determined experimentally, the computation times would easily exceed those reported for the MPL method. 

The main drawback of the MPL compared to the true ML is that the former in a sense over-specifies the dependence structure. As a result, the MPL is less data efficient than the ML, especially when the true network contains hub nodes. On the other hand, calculation of the true ML remains still intractable for non-chordal Markov networks. Therefore, the attractive properties of the MPL, combined with its solid performance in our experiments, suggest that our approach has considerable potential for both applications and further theoretical development.



\appendix
\section*{Appendix A.}

In this appendix we prove Theorem \ref{thm:MPLconsistency} from Section \ref{sec:MPL}:\\
\\
{\bf Theorem} {\it
Let $G^{*} \in\mathcal{G}$ be the true graph structure, of a Markov network over $(X_{1},\ldots,X_{d})$, with the corresponding Markov blankets $mb(G^{*})=\{ mb^{*}(1),\ldots,mb^{*}(d) \}$. Let $\theta_{G^{*}}\in\Theta_{G^{*}}$ define the corresponding joint distribution to which $G^{*}$ is faithful and from which a sample $\mathbf{x}$ of size $n$ is obtained.  The local MPL estimator
\begin{equation}\label{eq:locMPLest}
\skew{6}\widehat{m}b(j) = \underset{mb(j)\subseteq V\setminus j}{\arg \max} \ p(\mathbf{x}_{j} \mid \mathbf{x}_{mb(j)})
\end{equation} 
is consistent in the sense that $\skew{6}\widehat{m}b(j) = mb^{*}(j)$ eventually almost surely as $n\to \infty$ for $j=1,\ldots,d$. Consequently, the global MPL estimator
\begin{equation}\label{eq:globMPLest}
\hat G = \underset{G\in\mathcal{G}}{\arg \max} \ \hat p(\mathbf{x}\mid G)
\end{equation} 
is consistent in the sense that $\hat G = G^{*}$ eventually almost surely as $n\rightarrow \infty$.}\\
\\
\begin{proof}
The proof is based on an asymptotic comparison of the local log-MPL 
\begin{equation}\label{eq:local_MPL}
\begin{aligned}
\log  p(\mathbf{x}_{j} \mid \mathbf{x}_{mb(j)}) = & \sum_{l=1}^{q_{j}}[ \log \Gamma ( \alpha_{jl})-\log \Gamma ( n_{jl}+\alpha_{jl}) \\
+ & \sum_{i=1}^{r_{j}} \left( \log \Gamma ( n_{ijl}+\alpha_{ijl})-\log \Gamma ( \alpha_{ijl})\right)]
\end{aligned}
\end{equation}
and the PIC criterion \citep{Csiszar2006} which is defined by
\[
\text{PIC}(mb(j);\mathbf{x})=-\sum_{l=1}^{q_{j}}\sum_{i=1}^{r_{j}}n_{ijl}\log\frac{n_{ijl}}{n_{jl}}+q_{j}\log n 
\]
according to our notation. The PIC estimator is then defined as the Markov blanket (or neighborhood) that minimizes the above score. \citet{Csiszar2006} conclude that the PIC criterion is consistent (Theorem 2.1). Their proof is based on two key propositions (Proposition 4.1 and 5.1) which together rule out the possibility of overestimation as well as underestimation. They also note that their results remains valid if the penalty term is multiplied by any constant $c>0$. Thus, any estimator asymptotically equivalent to the PIC estimator is also consistent.

We proceed by investigating the asymptotic behavior of the local log-MPL when $n \to \infty$. We re-arrange the terms in \eqref{eq:local_MPL} and omit the constant term introducing an $O(1)$ error:
\begin{equation*}
\begin{aligned}
&\sum_{l=1}^{q_{j}}[ \log \Gamma ( \alpha_{jl})-\log \Gamma ( n_{jl}+\alpha_{jl})+ \sum_{i=1}^{r_{j}} \left( \log \Gamma ( n_{ijl}+\alpha_{ijl})-\log \Gamma ( \alpha_{ijl})\right)]\\
&=\sum_{l=1}^{q_{j}}[-\log \Gamma ( n_{jl}+\alpha_{jl})+ \sum_{i=1}^{r_{j}} \log \Gamma ( n_{ijl}+\alpha_{ijl})]+ \sum_{l=1}^{q_{j}}[\log \Gamma ( \alpha_{jl}) - \sum_{i=1}^{r_{j}} \log \Gamma ( \alpha_{ijl})]\\
&=\sum_{l=1}^{q_{j}}[-\log \Gamma ( n_{jl}+\alpha_{jl})+ \sum_{i=1}^{r_{j}} \log \Gamma ( n_{ijl}+\alpha_{ijl})]+ O(1)
\end{aligned}
\end{equation*}
We let $n\to\infty$ and apply Stirling's asymptotic formula
\begin{equation*}
\begin{aligned}
\log \Gamma(n)  	& \to (n-\frac{1}{2})\log n -n+O(1)
\end{aligned}
\end{equation*}
on the remaining terms:
\begin{equation*}
\begin{aligned}
&\sum_{l=1}^{q_{j}}[-\log \Gamma ( n_{jl}+\alpha_{jl})+ \sum_{i=1}^{r_{j}} \log \Gamma ( n_{ijl}+\alpha_{ijl})]+ O(1)\\
&\to\sum_{l=1}^{q_{j}}[-( n_{jl}+\alpha_{jl}-\frac{1}{2})\log( n_{jl}+\alpha_{jl})+( n_{jl}+\alpha_{jl})  \\
& + \sum_{i=1}^{r_{j}} ( n_{ijl}+\alpha_{ijl}-\frac{1}{2})\log( n_{ijl}+\alpha_{ijl})-( n_{ijl}+\alpha_{ijl})]+O(1) \\
& = \sum_{l=1}^{q_{j}}\sum_{i=1}^{r_{j}} n_{ijl} \log \frac{ n_{ijl}+\alpha_{ijl}}{ n_{jl}+\alpha_{jl}} + \sum_{l=1}^{q_{j}}\sum_{i=1}^{r_{j}}\alpha_{ijl}\log \frac{ n_{ijl}+\alpha_{ijl}}{ n_{jl}+\alpha_{jl}} \\
& + \sum_{l=1}^{q_{j}}[ \frac{1}{2}\log( n_{jl}+\alpha_{jl})-\sum_{i=1}^{r_{j}}\frac{1}{2}\log( n_{ijl}+\alpha_{ijl})]+ O(1)
\end{aligned}
\end{equation*}
The second step is allowed since $n_{jl}=\sum_{i=1}^{r_{j}}n_{ijl}$ and $\alpha_{jl}=\sum_{i=1}^{r_{j}}\alpha_{ijl}$. As $n\to\infty$ we have that
\[
\frac{ n_{ijl}+\alpha_{ijl}}{ n_{jl}+\alpha_{jl}} =\frac{n_{ijl}(1+\frac{\displaystyle{\alpha_{ijl}}}{\displaystyle{n_{ijl}}})}{n_{jl}(1+\frac{\displaystyle{\alpha_{jl}}}{\displaystyle{n_{jl}}})}\rightarrow \frac{n_{ijl}}{n_{jl}}
\]
Since $n_{ijl}/n_{jl}$ is the maximum likelihood estimate of the parameter $\theta_{ijl}$ we further know that
\[
\sum_{l=1}^{q_{j}}\sum_{i=1}^{r_{j}} \alpha_{ijl} \log \frac{ n_{ijl}+\alpha_{ijl}}{ n_{jl}+\alpha_{jl}}=O(1) 
\]
Finally, the remaining terms can be rewritten as
\begin{equation*}
\begin{aligned}
& \sum_{l=1}^{q_{j}}[ \frac{1}{2}\log( n_{jl}+\alpha_{jl})-\sum_{i=1}^{r_{j}}\frac{1}{2}\log( n_{ijl}+\alpha_{ijl})]\\
&=\frac{1}{2} \sum_{l=1}^{q_{j}}[ \log \frac{n_{jl}+\alpha_{jl}}{n} -\log n -\sum_{i=1}^{r_{j}}(\log \frac{n_{ijl}+\alpha_{ijl}}{n} -\log n )]\\
&=\frac{1}{2} \sum_{l=1}^{q_{j}}[-\log n +\sum_{i=1}^{r_{j}}\log n ]+\frac{1}{2} \sum_{l=1}^{q_{j}}[ \log \frac{n_{jl}+\alpha_{jl}}{n} -\sum_{i=1}^{r_{j}} \log \frac{n_{ijl}+\alpha_{ijl}}{n} ]\\
&=\frac{1}{2} \sum_{l=1}^{q_{j}}[-\log n +\sum_{i=1}^{r_{j}}\log n ]+O(1)\\
&=\frac{1}{2} (-q_{j}\log n +q_{j}r_{j}\log n )+O(1)\\
&=\frac{q_{j}(r_{j}-1)}{2} \log n +O(1)\\
\end{aligned}
\end{equation*}
Piecing everything together, 
\begin{equation*}
\log  p(\mathbf{x}_{j} \mid \mathbf{x}_{mb(j)}) \to \sum_{l=1}^{q_{j}}\sum_{i=1}^{r_{j}} n_{ijl} \log \frac{ n_{ijl}}{ n_{jl}}- \frac{(r_{j}-1)q_{j}}{2}\log n +O(1).
\end{equation*}
as $n\to\infty$. Since the $O(1)$ term does not grow with $n$, the local log-MPL is asymptotically equivalent to 
\begin{equation*}
\sum_{l=1}^{q_{j}}\sum_{i=1}^{r_{j}} n_{ijl} \log \frac{ n_{ijl}}{ n_{jl}}- c_{j}\cdot q_{j}\log n .
\end{equation*}
where $c_{j}=(r_{j}-1)/2$ is a variable specific constant. Consequently, the local MPL estimator is asymptotically equivalent to minimizing
\begin{equation*}
-\sum_{l=1}^{q_{j}}\sum_{i=1}^{r_{j}} n_{ijl} \log \frac{ n_{ijl}}{ n_{jl}}+c_{j}\cdot q_{j}\log n
\end{equation*}
which is equivalent to the consistent PIC estimator up to a constant factor on the penalty term. Hence the local MPL estimator \eqref{eq:locMPLest} is consistent.

Since the local MPL estimator is consistent, the true collection of Markov blankets is eventually identified when $n\to\infty$. A set of Markov blankets uniquely specifies the structure of a Markov network. Since the true model structure satisfies the structural properties of a Markov network, that is $i\in mb(j)$ if $j\in mb(i)$, the global MPL estimator \eqref{eq:globMPLest} is also consistent. 
\end{proof}

\section*{Appendix B.}
This appendix contains supplementary material of Section \ref{sec:Results} in form of numerical results.

\begin{sidewaystable}
\begin{centering}
{\footnotesize
\begin{tabular}{c|c|c| *{3}{c}|*{3}{c}}
\multirow{2}{*}{$d$} & \multirow{2}{*}{$n$} & Number of & \multicolumn{3}{c|}{PBO} & \multicolumn{3}{c}{HC}   \\
& & solved instances & Encoding/Solving time (s) & log-MPL & TP/FP & Time (s) & log-MPL &TP/FP\\
\hline
\hline
\multirow{8}{*}{64} &250 & 100 & 0.31/1.24 & -8700.20 & \textbf{34.16/6.23} & 0.11 & -8701.07 & 33.79/6.11 \\ 
&500 & 100 & 0.31/1.13 & -16974.31 & \textbf{41.12/2.97} & 0.13 & -16976.75 & 40.77/2.78 \\ 
&1000 & 100 & 0.26/0.67 & -33310.55 & \textbf{49.08/1.06} & 0.16 & -33317.19 & 48.55/1.08 \\ 
&2000 & 100 & 0.46/0.77 & -65613.39 & \textbf{55.75/0.46} & 0.25 & -65614.25 & 55.62/0.45 \\ 
&4000 & 100 & 0.70/0.76 & -132021.36 & 61.84/0.18 & 0.40 & -132021.37 & \textbf{61.85/0.17} \\ 
&8000 & 100 & 1.47/0.94 & -262370.09 & \textbf{65.33/0.13} & 0.74 & -262370.12 & \textbf{65.33/0.13} \\ 
&16000 & 100 & 1.57/0.49 & -519632.76 & \textbf{65.51/0.10} & 1.89 & -519632.76 & \textbf{65.51/0.10} \\ 
&32000 & 100 & 4.04/0.76 & -1015034.17 & 68.51/0.03 & 3.73 & -1015034.34 & \textbf{68.52/0.03} \\ 
\hline
\hline
\multirow{8}{*}{128} &250 & 91 & 4.42/43.63 & -17017.35 & 62.13/25.73 & 0.22 & -17021.41 & \textbf{61.62/23.91} \\ 
&500 & 100 & 1.22/7.19 & -33552.92 & \textbf{79.52/10.43} & 0.23 & -33561.81 & 78.45/10.22 \\ 
&1000 & 100 & 1.39/4.33 & -66074.15 & \textbf{97.25/4.42} & 0.29 & -66081.21 & 96.46/4.46 \\ 
&2000 & 100 & 0.92/2.05 & -131739.79 & \textbf{112.87/2.26} & 0.47 & -131742.17 & 112.68/2.22 \\ 
&4000 & 99 & 1.31/1.83 & -258402.88 & \textbf{122.91/0.94} & 0.76 & -258403.11 & 122.88/0.92 \\ 
&8000 & 100 & 2.18/1.71 & -514098.80 & \textbf{127.50/0.46} & 1.39 & -514098.95 & 127.45/0.46 \\ 
&16000 & 100 & 3.56/1.78 & -1030570.26 & \textbf{133.66/0.26} & 4.00 & -1030570.26 & \textbf{133.66/0.26} \\ 
&32000 & 100 & 8.77/2.09 & -2086638.66 & \textbf{137.16/0.09} & 7.81 & -2086639.27 & \textbf{137.16/0.09} \\ 
\hline
\hline
\multirow{8}{*}{256}&250 & 10 & 12.79/235.78 & -34505.12 & 120.70/64.00 & 0.48 & -34521.43 & \textbf{120.30/60.50} \\ 
&500 & 55 & 13.51/118.93 & -67470.66 & \textbf{161.13/31.95} & 0.49 & -67482.09 & 159.60/30.84 \\ 
&1000 & 90 & 6.26/28.85 & -132908.96 & \textbf{199.94/13.91} & 0.63 & -132927.70 & 198.66/13.52 \\ 
&2000 & 96 & 6.33/17.04 & -261032.09 & \textbf{222.43/7.21} & 0.92 & -261044.24 & 221.75/7.02 \\ 
&4000 & 100 & 4.65/7.33 & -518442.23 & \textbf{245.28/3.04} & 1.56 & -518446.10 & 245.17/2.99 \\ 
&8000 & 100 & 6.25/6.56 & -1031306.39 & \textbf{257.04/1.63} & 2.82 & -1031306.65 & \textbf{257.02/1.61} \\ 
&16000 & 100 & 8.53/4.22 & -2076927.13 & \textbf{267.76/0.47} & 8.13 & -2076927.30 & 267.70/0.47 \\ 
&32000 & 100 & 17.04/3.75 & -4075373.11 & \textbf{275.12/0.32} & 15.25 & -4075375.01 & 275.06/0.32 \\ 
\hline
\hline
\multirow{8}{*}{512}&250 & 0 & - & - & - & - & - & - \\ 
&500 & 0 & - & - & - & - & - & - \\ 
&1000 & 19 & 19.78/179.25 & -267311.61 & \textbf{379.37/42.00} & 1.37 & -267355.80 & 375.58/41.79 \\ 
&2000 & 44 & 9.24/44.10 & -522454.75 & \textbf{447.18/19.23} & 1.99 & -522464.46 & 446.55/18.84 \\ 
&4000 & 77 & 14.93/39.80 & -1043048.89 & \textbf{485.06/9.47} & 3.20 & -1043055.03 & 484.78/9.38 \\ 
&8000 & 97 & 15.08/19.84 & -2075735.29 & 513.11/4.18 & 5.84 & -2075735.89 & \textbf{513.11/4.14} \\ 
&16000 & 97 & 18.58/10.94 & -4166121.62 & \textbf{532.99/2.19} & 16.82 & -4166134.18 & 532.82/2.19 \\ 
&32000 & 100 & 34.25/8.71 & -8296145.22 & \textbf{553.44/1.06} & 31.54 & -8296146.02 & 553.38/1.06 \\ 
\end{tabular}
}
\caption{Results from the comparison of the PBO and HC method in Section \ref{sec:SYNTH_SIM}. (\textbf{bold font} $=$ lowest Hamming distance) \label{tab:PBO_HC_COMP}}
\end{centering}
\end{sidewaystable}

\begin{sidewaystable}
\begin{centering}
{\footnotesize
\begin{tabular}{c|c|c|c|*{2}{c}|*{2}{c}}
\multirow{2}{*}{$d$} & \multirow{2}{*}{$n$} & \multirow{2}{*}{MPL} & \multirow{2}{*}{PIC} & \multicolumn{2}{c}{CMI} & \multicolumn{2}{c}{L1LR} \\
& & & & AND & OR & AND & OR \\
\hline
\hline
\multirow{8}{*}{64} &250 & \textbf{33.79/6.11} & 23.77/0.34 & 22.74/1.29 & 23.22/2.36 & 24.11/1.74 & 26.96/2.26 \\ 
&500 & \textbf{40.77/2.78} & 29.43/0.09 & 29.03/0.45 & 24.44/1.58 & 32.31/3.23 & 34.38/3.31 \\ 
&1000 & \textbf{48.55/1.08} & 39.09/0.07 & 41.65/5.87 & 37.57/3.20 & 39.86/1.69 & 44.54/2.86 \\ 
&2000 & \textbf{55.62/0.45} & 46.81/0.02 & 49.18/1.30 & 37.37/0.12 & 47.41/1.99 & 50.50/3.65 \\ 
&4000 & \textbf{61.85/0.17} & 55.90/0.01 & 57.16/0.49 & 58.98/4.66 & 55.66/1.19 & 59.62/2.52 \\ 
&8000 & \textbf{65.33/0.13} & 60.92/0.01 & 62.70/1.59 & 57.32/2.20 & 62.51/1.41 & 65.20/2.96 \\ 
&16000 & \textbf{65.51/0.10} & 61.84/0.00 & 61.42/0.58 & 59.29/1.51 & 60.23/0.72 & 63.52/2.22 \\ 
&32000 & \textbf{68.52/0.03} & 65.47/0.00 & 68.45/0.03 & 58.61/0.87 & 66.13/0.84 & 66.60/1.92 \\ 
\hline
\hline
\multirow{8}{*}{128} &250 & 61.09/24.04 & \textbf{44.28/1.85} & 38.10/0.54 & 37.33/2.82 & 40.48/3.06 & 46.38/6.39 \\ 
&500 & \textbf{78.45/10.22} & 58.32/0.51 & 57.15/0.73 & 46.68/1.14 & 59.79/5.08 & 67.32/11.16 \\ 
&1000 & \textbf{96.46/4.46} & 75.49/0.20 & 81.98/11.04 & 69.99/10.80 & 73.98/2.51 & 82.76/7.35 \\ 
&2000 & \textbf{112.68/2.22} & 97.49/0.04 & 97.53/3.25 & 73.20/0.64 & 95.28/4.43 & 101.51/11.33 \\ 
&4000 & \textbf{122.88/0.93} & 111.89/0.06 & 108.58/1.87 & 114.37/9.46 & 105.43/2.44 & 111.84/6.69 \\ 
&8000 & \textbf{127.45/0.46} & 118.91/0.00 & 121.44/2.99 & 117.22/8.50 & 116.72/3.78 & 117.89/5.81 \\ 
&16000 & \textbf{133.66/0.26} & 126.10/0.01 & 125.24/0.25 & 119.42/3.72 & 124.84/2.22 & 127.07/4.28 \\ 
&32000 & \textbf{137.16/0.09} & 130.81/0.00 & 136.62/0.18 & 122.34/1.35 & 131.61/2.84 & 130.49/2.15 \\ 
\hline
\hline
\multirow{8}{*}{256} &250 & 118.30/66.51 & \textbf{87.70/6.81} & 75.94/0.79 & 68.19/1.55 & 81.55/12.82 & 90.90/21.09 \\ 
&500 & \textbf{158.28/32.72} & 119.64/1.98 & 117.51/2.05 & 93.90/1.36 & 123.63/20.07 & 127.14/28.66 \\ 
&1000 & \textbf{198.98/13.76} & 157.33/0.53 & 168.82/22.57 & 146.36/32.54 & 150.65/9.98 & 163.52/19.68 \\ 
&2000 & \textbf{221.73/7.08} & 188.35/0.22 & 191.71/5.95 & 145.50/2.03 & 183.54/15.62 & 175.56/10.07 \\ 
&4000 & \textbf{245.17/2.99} & 222.17/0.05 & 223.00/3.08 & 225.08/28.10 & 210.33/9.00 & 210.06/7.13 \\ 
&8000 & \textbf{257.02/1.61} & 239.52/0.03 & 242.47/5.14 & 225.79/17.45 & 236.47/13.77 & 227.10/4.39 \\ 
&16000 & \textbf{267.70/0.47} & 255.05/0.01 & 253.66/0.27 & 237.94/7.68 & 253.64/8.56 & 248.94/6.57 \\ 
&32000 & \textbf{275.06/0.32} & 264.72/0.00 & 271.74/0.07 & 243.56/2.18 & 262.54/9.02 & 253.13/5.90 \\ 
\hline
\hline
\multirow{8}{*}{512} &250 & 238.83/168.39 & \textbf{182.13/22.64} & 159.25/2.77 & 145.26/5.19 & 160.15/37.21 & 134.20/4.10 \\ 
&500 & 303.82/85.51 & \textbf{235.79/7.33} & 228.66/4.57 & 189.02/7.22 & 206.20/40.99 & 187.37/2.49 \\ 
&1000 & \textbf{384.26/40.57} & 306.16/2.05 & 333.11/35.84 & 282.97/74.35 & 298.57/38.14 & 276.56/6.36 \\ 
&2000 & \textbf{442.16/20.10} & 377.77/0.63 & 374.03/8.48 & 287.87/3.36 & 345.15/42.16 & 325.34/3.19 \\ 
&4000 & \textbf{484.44/9.65} & 439.55/0.17 & 432.42/4.06 & 432.73/60.98 & 417.76/30.00 & 406.06/9.57 \\ 
&8000 & \textbf{513.00/4.17} & 476.57/0.04 & 485.44/7.56 & 442.44/33.06 & 433.84/12.07 & 434.82/14.64 \\ 
&16000 & \textbf{532.61/2.25} & 504.79/0.04 & 502.86/0.25 & 470.29/23.79 & 489.49/15.26 & 447.91/15.35 \\ 
&32000 & \textbf{553.38/1.06} & 529.72/0.02 & 545.69/0.12 & 488.06/6.44 & 505.80/1.25 & 457.72/14.94 \\ 
\end{tabular}
}
\caption{True and false positives (TP/FP) for the methods used in Section \ref{sec:SYNTH_SIM}. (\textbf{bold font} $=$ lowest Hamming distance) \label{tab:COMP_TPFP}}
\end{centering}
\end{sidewaystable}

\begin{sidewaystable}
\begin{centering}
{\footnotesize
\begin{tabular}{c|c|c|c|*{2}{c}|*{2}{c}|c|c}
\multirow{3}{*}{$d$} & \multirow{3}{*}{$n$} & \multicolumn{6}{c|}{Markov blanket discovery  (total/node max)} & \multicolumn{2}{c}{Hill climbing} \\
& & \multirow{2}{*}{MPL} & \multirow{2}{*}{PIC} & \multicolumn{2}{c|}{CMI} & \multicolumn{2}{c|}{L1LR} & \multirow{2}{*}{MPL} & \multirow{2}{*}{PIC} \\
& & & & AND & OR & AND & OR & & \\
\hline
\hline
\multirow{8}{*}{64}&250 & 1.83/0.11 & 1.25/0.06 & 1.17/0.11 & 0.52/0.05 & 0.96/0.05 & 0.97/0.05 & 0.11 & 0.07 \\ 
&500 & 2.18/0.13 & 1.61/0.08 & 1.76/0.25 & 0.61/0.06 & 1.23/0.06 & 1.26/0.06 & 0.13 & 0.09 \\ 
&1000 & 2.93/0.15 & 2.42/0.11 & 7.93/0.52 & 1.08/0.16 & 1.66/0.08 & 1.65/0.08 & 0.16 & 0.12 \\ 
&2000 & 4.83/0.25 & 4.19/0.21 & 10.08/1.23 & 1.42/0.10 & 3.51/0.16 & 3.48/0.16 & 0.25 & 0.20 \\ 
&4000 & 8.40/0.39 & 7.70/0.33 & 18.26/2.60 & 4.56/0.91 & 9.97/0.46 & 9.86/0.46 & 0.40 & 0.35 \\ 
&8000 & 15.80/0.85 & 14.81/0.64 & 55.00/7.12 & 7.70/1.19 & 24.81/1.02 & 23.76/1.00 & 0.74 & 0.66 \\ 
&16000 & 45.14/2.20 & 42.34/1.82 & 65.86/13.39 & 18.19/2.33 & 69.83/3.05 & 67.39/2.99 & 1.89 & 1.38 \\ 
&32000 & 84.43/4.96 & 78.33/3.81 & 147.63/40.20 & 43.89/4.52 & 164.17/6.71 & 148.07/6.09 & 3.73 & 2.81 \\
\hline
\hline
\multirow{8}{*}{128}&250 & 8.45/0.24 & 4.78/0.10 & 2.97/0.24 & 1.82/0.09 & 2.84/0.06 & 2.84/0.06 & 0.22 & 0.10 \\ 
&500 & 9.02/0.24 & 6.39/0.16 & 8.03/0.54 & 2.31/0.12 & 3.75/0.08 & 3.74/0.08 & 0.23 & 0.14 \\ 
&1000 & 12.14/0.32 & 9.45/0.24 & 42.90/1.20 & 4.57/0.58 & 6.57/0.14 & 6.57/0.14 & 0.29 & 0.20 \\ 
&2000 & 20.30/0.49 & 17.57/0.46 & 47.29/2.39 & 5.75/0.28 & 18.63/0.42 & 18.43/0.41 & 0.47 & 0.38 \\ 
&4000 & 33.77/0.81 & 30.61/0.68 & 78.82/5.24 & 17.21/2.29 & 37.88/0.80 & 36.67/0.78 & 0.76 & 0.66 \\ 
&8000 & 62.61/1.74 & 58.88/1.43 & 209.36/13.22 & 34.73/5.71 & 117.10/2.27 & 105.14/2.10 & 1.39 & 1.21 \\ 
&16000 & 186.96/5.18 & 174.55/3.96 & 160.95/26.09 & 74.74/8.73 & 337.29/7.65 & 294.69/6.64 & 4.00 & 2.86 \\ 
&32000 & 343.98/11.83 & 323.12/9.15 & 510.09/82.21 & 185.69/11.66 & 829.72/19.47 & 668.22/15.11 & 7.81 & 5.60 \\
\hline
\hline
\multirow{8}{*}{256}&250 & 40.96/0.67 & 19.63/0.22 & 13.21/0.55 & 6.77/0.16 & 8.61/0.09 & 8.66/0.09 & 0.54 & 0.17 \\ 
&500 & 41.38/0.60 & 26.24/0.32 & 41.37/1.11 & 9.19/0.23 & 15.18/0.17 & 15.56/0.17 & 0.50 & 0.24 \\ 
&1000 & 52.61/0.72 & 39.49/0.58 & 218.80/2.40 & 20.05/1.45 & 41.89/0.44 & 41.60/0.44 & 0.63 & 0.38 \\ 
&2000 & 80.98/1.10 & 67.61/0.90 & 222.70/4.80 & 22.87/0.95 & 99.96/1.03 & 88.52/0.94 & 0.92 & 0.67 \\ 
&4000 & 138.11/2.08 & 125.06/1.39 & 342.81/10.50 & 69.60/6.18 & 218.26/2.41 & 186.89/2.02 & 1.56 & 1.29 \\ 
&8000 & 256.81/4.36 & 238.05/3.13 & 959.13/27.50 & 125.43/14.78 & 573.22/7.45 & 406.35/5.13 & 2.82 & 2.44 \\ 
&16000 & 755.30/11.57 & 710.80/9.12 & 651.17/57.73 & 284.78/26.34 & 1536.93/21.77 & 1110.54/14.63 & 8.13 & 5.74 \\ 
&32000 & 1404.29/27.84 & 1299.78/19.75 & 1996.34/171.50 & 719.29/31.50 & 3150.53/41.52 & 1948.53/26.06 & 15.25 & 10.97 \\
\hline
\hline
\multirow{8}{*}{512}&250 & 208.35/2.16 & 83.19/0.48 & 67.56/1.17 & 28.05/0.51 & 39.11/0.22 & 43.20/0.23 & 1.59 & 0.37 \\ 
&500 & 183.72/1.33 & 105.99/0.65 & 193.05/2.27 & 37.88/0.99 & 90.13/0.47 & 93.68/0.46 & 1.24 & 0.49 \\ 
&1000 & 220.51/1.53 & 156.32/1.25 & 1001.81/4.73 & 81.83/3.12 & 192.60/1.03 & 182.68/0.92 & 1.41 & 0.76 \\ 
&2000 & 334.71/2.41 & 272.74/1.80 & 1039.68/9.91 & 89.86/2.44 & 443.78/2.78 & 369.68/1.96 & 2.01 & 1.37 \\ 
&4000 & 557.18/4.62 & 494.41/2.82 & 1445.58/21.49 & 276.06/14.89 & 1131.22/7.96 & 862.90/5.38 & 3.20 & 2.54 \\ 
&8000 & 1031.36/9.70 & 957.13/6.93 & 4374.04/55.34 & 472.23/31.79 & 2404.33/17.21 & 1925.95/13.67 & 5.85 & 4.95 \\ 
&16000 & 3058.00/29.74 & 2842.66/19.69 & 2261.52/111.66 & 1057.63/71.35 & 7002.39/45.74 & 3614.59/27.36 & 16.81 & 11.28 \\ 
&32000 & 5697.65/64.94 & 5337.01/44.28 & 8494.37/355.76 & 2782.49/98.06 & 16049.32/92.76 & 8041.91/59.57 & 31.54 & 22.68 \\ 
\end{tabular}
}
\caption{Execution times (in seconds) for the methods used in Section \ref{sec:SYNTH_SIM}. \label{tab:COMP_TIMES}}
\end{centering}
\end{sidewaystable}

\begin{sidewaystable}
\begin{centering}
{\footnotesize
\begin{subtable}[]{0.59\textwidth}
\begin{tabular}{c|c|c|c|*{2}{c}}
\multirow{2}{*}{Network} & \multirow{2}{*}{$n$} & \multirow{2}{*}{MPL} & \multirow{2}{*}{PIC} & \multicolumn{2}{c}{CMI}  \\
& & & & AND & OR  \\
\hline
\hline
\multirow{8}{*}{Alarm}&250 & \textbf{38.48/3.70} & 17.87/1.87 & 19.55/2.53 &  30.73/5.71  \\ 
&500 & \textbf{43.05/1.39} & 19.96/2.04 & 22.48/0.93 &  33.51/3.00  \\ 
&1000 & \textbf{46.22/0.39} & 25.62/2.02 & 28.15/2.73 & 37.60/3.92  \\ 
&2000 & \textbf{49.02/0.37} & 31.59/1.40 & 30.37/0.06 &  39.39/2.38  \\ 
&4000 & \textbf{50.65/0.25} & 34.94/0.99 & 34.44/2.00 & 42.22/2.63 \\ 
&8000 & \textbf{52.40/0.35} & 39.00/0.34 & 39.76/0.50 & 47.28/1.95 \\ 
&16000 & \textbf{53.98/0.74} & 42.77/0.21 & 43.78/0.27 &  49.57/2.02 \\ 
&32000 & \textbf{55.97/0.98} & 44.00/1.02 & 45.45/0.57 & 53.98/1.58 \\ 
\hline
\hline
\multirow{8}{*}{Insurance}&250 & \textbf{21.75/1.95} & 14.37/0.69 & 15.67/3.02 &  18.31/2.75  \\ 
&500 & \textbf{25.26/1.44} & 16.82/0.93 & 18.75/1.41 & 30.26/10.90  \\ 
&1000 & \textbf{28.95/1.02} & 20.21/1.11 & 21.06/1.54 & 32.98/8.22 \\ 
&2000 & \textbf{32.04/0.82} & 22.12/0.97 & 22.97/2.81 & 31.95/4.61  \\ 
&4000 & \textbf{34.94/0.78} & 26.03/0.93 & 27.62/2.91 & 33.32/2.56 \\ 
&8000 & \textbf{38.67/0.87} & 28.63/0.92 & 31.91/2.40 & 37.09/4.52  \\ 
&16000 & \textbf{40.87/0.92} & 31.02/0.88 & 32.52/1.05 & 37.85/1.03 \\ 
&32000 & \textbf{43.61/0.82} & 34.00/0.99 & 40.63/2.32  & 36.99/0.99  \\ 
\hline
\hline
\multirow{8}{*}{Hailfinder}&250 & 25.83/8.97 & \textbf{21.52/3.34} & 16.28/2.58 &  18.18/4.65 \\ 
&500 & 30.83/9.25 & \textbf{25.50/3.01} & 16.92/3.54  & 20.14/2.88 \\ 
&1000 & 36.08/10.31 & 30.90/2.19 & 25.28/6.85  & \textbf{35.50/1.66} \\ 
&2000 & 38.60/10.64 & 35.45/2.65 & 23.16/3.54  & \textbf{34.95/1.31}  \\ 
&4000 & 43.68/9.54 & 38.69/3.20 & 28.05/0.23  & \textbf{49.81/4.59} \\ 
&8000 & 52.09/9.42 & 43.05/2.81 & 34.66/0.02 & \textbf{50.41/4.82}  \\ 
&16000 & 55.71/8.69 & 47.57/2.78 & 35.75/0.17  & \textbf{52.38/5.25}  \\ 
&32000 & \textbf{59.21/8.84} & 50.68/3.89 & 37.01/0.00 & 52.84/4.89  \\ 
\hline
\hline
\multirow{8}{*}{Barley}&250 & \textbf{16.69/3.93} & 13.52/2.39 & 3.25/0.67 & 4.90/1.73 \\ 
&500 & 19.86/3.61 & \textbf{20.64/2.99} & 9.55/3.17 & 14.70/6.15  \\ 
&1000 & \textbf{27.28/3.18} & 24.84/2.99 & 16.28/4.37 & 22.59/8.69  \\ 
&2000 & \textbf{32.67/3.45} & 28.89/3.02 & 17.65/5.78 & 22.98/6.00  \\ 
&4000 & \textbf{39.68/3.25} & 30.53/3.00 & 19.37/5.15 & 26.26/6.16  \\ 
&8000 & \textbf{44.03/3.39} & 34.18/3.00 & 17.52/2.00  & 40.63/13.47  \\ 
&16000 & \textbf{49.49/3.03} & 39.40/3.21 & 19.27/2.60 &  37.68/14.61  \\ 
&32000 & \textbf{54.95/3.09} & 39.71/3.00 & 24.67/2.13  & 42.00/8.00 \\ 
\end{tabular}
\caption{\label{tab:BM_TPFP}}
\end{subtable}
\begin{subtable}[]{0.35\textwidth}
\begin{tabular}{c|c|c}
\multirow{2}{*}{$n$} & \multirow{2}{*}{$N$} &\multirow{2}{*}{MPL}  \\
& & \\
\hline
\hline
250 & \multirow{8}{*}{1} & \textbf{38.48/3.70} \\ 
500 & & \textbf{43.05/1.39} \\ 
1000 & & 46.22/0.39 \\ 
2000 & & 49.02/0.37 \\ 
4000 & & \textbf{50.65/0.25} \\ 
8000 & & 52.40/0.35 \\ 
16000 & & 53.98/0.74 \\ 
32000 & & 55.97/0.98 \\
\hline
\hline
250 & \multirow{8}{*}{4} & 40.77/6.49 \\ 
500 & & 44.75/3.16 \\ 
1000 & & \textbf{47.39/1.29} \\ 
2000 & & \textbf{49.48/0.75} \\ 
4000 & & \textbf{50.85/0.45} \\ 
8000 & & \textbf{53.01/0.83} \\ 
16000 & & 54.20/0.98 \\ 
32000 & & 56.26/1.01 \\  
\hline
\hline
250 & \multirow{8}{*}{8} & 42.33/10.85 \\ 
500 & & 45.76/6.28 \\ 
1000 & & 48.16/3.65 \\ 
2000 & & 50.03/2.51 \\ 
4000 & & 51.30/1.95 \\ 
8000 & & 53.38/1.70 \\ 
16000 & & \textbf{54.80/1.33} \\ 
32000 & & \textbf{56.46/1.11} \\ 
\hline
\hline
250 & \multirow{8}{*}{16} & 44.11/17.37 \\ 
500 & & 47.08/12.02 \\ 
1000 & & 49.24/8.28 \\ 
2000 & & 50.69/5.91 \\ 
4000 & & 52.12/4.10 \\ 
8000 & & 53.93/3.31 \\ 
16000 & & 55.47/2.78 \\ 
32000 & & 56.70/2.12 \\ 
\end{tabular}
\begin{tabular}{c|c}
 \multirow{2}{*}{$N$} &\multirow{2}{*}{MPL}  \\
& \\
\hline
\hline
\multirow{8}{*}{32}& 45.38/25.52 \\ 
& 48.16/19.47 \\ 
& 50.31/14.92 \\ 
& 51.33/11.72 \\ 
& 53.04/8.53 \\ 
& 54.40/7.08 \\ 
& 55.78/5.30 \\ 
& 57.39/4.02 \\ 
\hline
\hline
\multirow{8}{*}{64}& 45.96/35.34 \\ 
& 48.67/29.15 \\ 
& 50.67/24.12 \\ 
& 51.92/19.47 \\ 
& 53.95/15.30 \\ 
& 55.38/12.99 \\ 
& 56.28/10.42 \\ 
& 57.25/8.19 \\ 
\hline
\hline
\multirow{8}{*}{128}& 46.40/47.58 \\ 
& 49.27/40.37 \\ 
& 50.72/34.63 \\ 
& 52.49/29.80 \\ 
& 54.63/24.74 \\ 
& 55.96/20.65 \\ 
& 56.58/17.48 \\ 
& 57.20/14.91 \\ 
\hline
\hline
\multirow{8}{*}{256}& 47.00/59.48 \\ 
& 49.79/52.86 \\ 
& 51.69/46.52 \\ 
& 52.78/41.55 \\ 
& 54.71/36.27 \\ 
& 56.12/31.58 \\ 
& 56.76/27.60 \\ 
& 57.55/24.35 \\ 
\end{tabular}
\caption{\label{tab:ESS_TPFP}}
\end{subtable}
}
\caption{True and false positives (TP/FP) for the simulations in Section \ref{sec:BM_SIM} for (a) the methods used in the comparison, and (b) the MPL-HC applied on the Alarm network under different values of the equivalent sample size parameter $N$. (\textbf{bold font} $=$ lowest Hamming distance) \label{tab:BM_TPFP}}
\end{centering}
\end{sidewaystable}

\begin{sidewaystable}
\begin{centering}
{\footnotesize
\begin{tabular}{c|c|c|c|*{2}{c}|c|c}
\multirow{3}{*}{Network} & \multirow{3}{*}{$n$} & \multicolumn{4}{c|}{Markov blanket discovery  (total/node max)} & \multicolumn{2}{c}{Hill climbing} \\
& & \multirow{2}{*}{MPL} & \multirow{2}{*}{PIC} & \multicolumn{2}{c|}{CMI} & \multirow{2}{*}{MPL} & \multirow{2}{*}{PIC} \\
& & & & AND & OR & & \\
\hline
\hline
\multirow{8}{*}{Alarm}&250 & 1.31/0.09 & 0.67/0.04 & 0.43/0.06 & 0.28/0.03 & 0.18 & 0.08 \\ 
&500 & 1.71/0.12 & 0.94/0.04 & 0.53/0.08 & 0.33/0.02 & 0.23 & 0.10 \\ 
&1000 & 2.42/0.19 & 1.38/0.06 & 1.15/0.27 & 0.50/0.05 & 0.33 & 0.13 \\ 
&2000 & 3.96/0.35 & 2.32/0.13 & 1.34/0.38 & 0.75/0.06 & 0.54 & 0.21 \\ 
&4000 & 7.21/0.73 & 4.08/0.22 & 4.18/1.08 & 1.53/0.19 & 0.98 & 0.35 \\ 
&8000 & 14.92/1.69 & 8.28/0.43 & 9.25/3.47 & 3.28/0.22 & 2.09 & 0.74 \\ 
&16000 & 37.84/4.52 & 23.56/1.51 & 23.52/8.39 & 7.58/0.50 & 5.32 & 1.98 \\ 
&32000 & 78.17/10.79 & 47.98/3.31 & 59.65/21.41 & 22.09/1.53 & 11.48 & 3.99 \\ 
\hline
\hline
\multirow{8}{*}{Insurance}&250 & 0.64/0.06 & 0.45/0.04 & 0.33/0.04 & 0.13/0.01 & 0.11 & 0.07 \\ 
&500 & 0.89/0.08 & 0.62/0.04 & 0.46/0.06 & 0.29/0.05 & 0.13 & 0.09 \\ 
&1000 & 1.37/0.12 & 0.93/0.05 & 0.75/0.13 & 0.40/0.08 & 0.21 & 0.13 \\ 
&2000 & 2.36/0.20 & 1.61/0.14 & 1.51/0.28 & 0.50/0.10 & 0.40 & 0.20 \\ 
&4000 & 4.36/0.58 & 2.99/0.25 & 3.52/0.78 & 0.83/0.15 & 0.73 & 0.43 \\ 
&8000 & 8.94/1.27 & 5.85/0.47 & 6.82/1.57 & 2.13/0.67 & 1.64 & 0.79 \\ 
&16000 & 25.27/3.26 & 15.28/1.62 & 12.66/3.24 & 3.85/0.59 & 4.43 & 1.94 \\ 
&32000 & 57.61/11.37 & 35.42/4.91 & 55.46/15.08 & 9.45/1.16 & 10.69 & 4.86 \\ 
\hline
\hline
\multirow{8}{*}{Hailfinder}&250 & 3.02/0.19 & 2.68/0.09 & 1.64/0.05 & 0.37/0.03 & 0.19 & 0.12 \\ 
&500 & 4.13/0.22 & 3.69/0.13 & 2.00/0.10 & 0.43/0.03 & 0.31 & 0.16 \\ 
&1000 & 5.94/0.30 & 5.19/0.19 & 4.61/0.20 & 0.66/0.03 & 0.65 & 0.24 \\ 
&2000 & 8.98/0.41 & 8.98/0.44 & 4.84/0.43 & 0.89/0.04 & 1.44 & 0.38 \\ 
&4000 & 15.02/0.75 & 15.71/0.96 & 5.77/0.79 & 2.34/0.12 & 3.71 & 0.64 \\ 
&8000 & 32.41/2.16 & 28.81/1.63 & 18.35/1.92 & 4.74/0.36 & 13.24 & 1.19 \\ 
&16000 & 82.42/6.04 & 68.38/5.11 & 39.27/4.99 & 10.72/0.95 & 35.46 & 2.84 \\ 
&32000 & 179.16/22.53 & 152.45/17.70 & 47.21/9.80 & 26.04/1.75 & 124.36 & 6.20 \\ 
\hline
\hline
\multirow{8}{*}{Barley}&250 & 3.64/0.28 & 2.80/0.15 & 0.28/0.03 & 0.22/0.02 & 0.13 & 0.10 \\ 
&500 & 6.92/0.68 & 5.93/0.29 & 1.15/0.08 & 0.44/0.06 & 0.20 & 0.18 \\ 
&1000 & 11.20/1.02 & 10.56/0.51 & 3.45/0.23 & 0.75/0.15 & 0.37 & 0.31 \\ 
&2000 & 19.93/2.31 & 18.70/1.13 & 7.94/0.47 & 0.91/0.24 & 0.73 & 0.52 \\ 
&4000 & 50.08/6.39 & 34.09/1.92 & 12.24/1.08 & 1.89/0.66 & 2.12 & 1.11 \\ 
&8000 & 117.97/13.93 & 71.36/4.79 & 7.46/2.70 & 7.46/2.70 & 4.52 & 2.22 \\ 
&16000 & 346.72/47.56 & 155.19/10.12 & 26.38/5.73 & 16.36/4.72 & 11.43 & 5.45 \\ 
&32000 & 921.36/173.24 & 357.36/26.95 & 103.51/16.75 & 33.59/16.90 & 36.83 & 12.74 \\ \ 
\end{tabular}
}
\caption{Execution times (in seconds) for the methods used in Section \ref{sec:BM_SIM}. \label{tab:BM_TIMES}}
\end{centering}
\end{sidewaystable}

\begin{sidewaystable}
\begin{centering}
{\footnotesize
\begin{subtable}[]{0.55\textwidth}
\begin{tabular}{c|c|c c|*{2}{c}}
\multirow{2}{*}{$d$} & \multirow{2}{*}{$n$} &  \multicolumn{2}{c|}{CMI} & \multicolumn{2}{c}{L1LR}  \\
& & AND & OR & AND & OR  \\
\hline
\hline
\multirow{8}{*}{64}&250 & 0.02500/0.05000 & 0.05000/0.10000 & 8/16 & 12/16 \\ 
&500 & 0.02500/0.02500 & 0.02500/0.05000 & 12/16 & 16/24 \\ 
&1000 & 0.01000/0.02500 & 0.02500/0.05000 & 16/24 & 24/32 \\ 
&2000 & 0.00500/0.01000 & 0.01000/0.02500 & 24/32 & 24/48 \\ 
&4000 & 0.00500/0.00750 & 0.00750/0.01000 & 32/48 & 48/64 \\ 
&8000 & 0.00250/0.00500 & 0.00500/0.01000 & 48/64 & 64/96 \\ 
&16000 & 0.00100/0.00250 & 0.00250/0.01000 & 64/128 & 96/128 \\ 
&32000 & 0.00075/0.00100 & 0.00100/0.00750 & 128/192 & 128/256 \\ 
\hline
\hline
\multirow{8}{*}{128}&250 & 0.02500/0.05000 & 0.05000/0.07500 & 12/12 & 12/16 \\ 
&500 & 0.02500/0.02500 & 0.05000/0.07500 & 16/16 & 16/24 \\ 
&1000 & 0.01000/0.02500 & 0.02500/0.05000 & 24/24 & 24/32 \\ 
&2000 & 0.00500/0.01000 & 0.02500/0.02500 & 32/32 & 32/48 \\ 
&4000 & 0.00500/0.00750 & 0.00750/0.01000 & 48/48 & 48/64 \\ 
&8000 & 0.00250/0.00500 & 0.00500/0.01000 & 64/96 & 64/96 \\ 
&16000 & 0.00100/0.00250 & 0.00250/0.01000 & 96/128 & 96/128 \\ 
&32000 & 0.00075/0.00100 & 0.00250/0.00750 & 128/192 & 128/256 \\ 
\hline
\hline
\multirow{8}{*}{256}&250 & 0.05000/0.05000 & 0.07500/0.07500 & 12/16 & 12/16 \\ 
&500 & 0.02500/0.02500 & 0.05000/0.05000 & 16/16 & 16/24 \\ 
&1000 & 0.01000/0.02500 & 0.02500/0.05000 & 24/24 & 24/32 \\ 
&2000 & 0.00500/0.01000 & 0.02500/0.02500 & 32/32 & 32/48 \\ 
&4000 & 0.00500/0.00750 & 0.00750/0.01000 & 48/48 & 48/64 \\ 
&8000 & 0.00250/0.00500 & 0.00500/0.01000 & 64/96 & 64/128 \\ 
&16000 & 0.00100/0.00250 & 0.00500/0.01000 & 96/128 & 96/192 \\ 
&32000 & 0.00075/0.00100 & 0.00250/0.00750 & 128/192 & 192/512 \\ 
\hline
\hline
\multirow{8}{*}{512}&250 & 0.05000/0.05000 & 0.07500/0.07500 & 12/16 & 12/16 \\ 
&500 & 0.02500/0.02500 & 0.05000/0.05000 & 16/24 & 24/24 \\ 
&1000 & 0.01000/0.01000 & 0.02500/0.05000 & 24/32 & 24/32 \\ 
&2000 & 0.00500/0.01000 & 0.02500/0.02500 & 32/48 & 48/64 \\ 
&4000 & 0.00500/0.00500 & 0.01000/0.01000 & 48/64 & 64/96 \\ 
&8000 & 0.00250/0.00250 & 0.00750/0.01000 & 64/96 & 96/128 \\ 
&16000 & 0.00100/0.00250 & 0.00500/0.01000 & 96/128 & 128/256 \\ 
&32000 & 0.00100/0.00100 & 0.00500/0.00750 & 192/192 & 256/512 \\ 
\end{tabular}
\caption{}
\end{subtable}
\begin{subtable}[]{0.44\textwidth}
\begin{tabular}{c|c|c c}
\multirow{2}{*}{Network} & \multirow{2}{*}{$n$} & \multicolumn{2}{c}{CMI}  \\
& & AND & OR  \\
\hline
\hline
\multirow{8}{*}{Alarm}&250 & 0.05000/0.10000 & 0.07500/0.25000 \\ 
&500 & 0.02500/0.07500 & 0.05000/0.10000 \\ 
&1000 & 0.02500/0.05000 & 0.05000/0.10000 \\ 
&2000 & 0.02500/0.02500 & 0.02500/0.07500 \\ 
&4000 & 0.01000/0.02500 & 0.02500/0.05000 \\ 
&8000 & 0.00750/0.01000 & 0.02500/0.02500 \\ 
&16000 & 0.00500/0.01000 & 0.01000/0.02500 \\ 
&32000 & 0.00250/0.00750 & 0.00750/0.01000 \\
\hline
\hline
\multirow{8}{*}{Insurance}&250 & 0.05000/0.25000 & 0.07500/0.25000 \\ 
&500 & 0.05000/0.10000 & 0.07500/0.25000 \\ 
&1000 & 0.05000/0.10000 & 0.05000/0.10000 \\ 
&2000 & 0.02500/0.07500 & 0.05000/0.10000 \\ 
&4000 & 0.02500/0.05000 & 0.05000/0.10000 \\ 
&8000 & 0.01000/0.02500 & 0.05000/0.07500 \\ 
&16000 & 0.00750/0.02500 & 0.05000/0.05000 \\ 
&32000 & 0.00750/0.01000 & 0.05000/0.05000 \\
\hline
\hline
\multirow{8}{*}{Hailfinder}&250 & 0.05000/0.25000 & 0.50000/0.75000 \\ 
&500 & 0.02500/0.25000 & 0.25000/0.50000 \\ 
&1000 & 0.02500/0.10000 & 0.25000/0.25000 \\ 
&2000 & 0.05000/0.10000 & 0.25000/0.25000 \\ 
&4000 & 0.07500/0.10000 & 0.10000/0.10000 \\ 
&8000 & 0.05000/0.05000 & 0.07500/0.10000 \\ 
&16000 & 0.02500/0.05000 & 0.07500/0.10000 \\ 
&32000 & 0.05000/0.05000 & 0.05000/0.10000 \\ 
\hline
\hline
\multirow{8}{*}{Barley}&250 & 0.10000/2.00000 & 1.25000/2.00000 \\ 
&500 & 0.01000/1.25000 & 1.00000/1.25000 \\ 
&1000 & 0.00500/0.50000 & 0.75000/1.00000 \\ 
&2000 & 0.00500/0.25000 & 0.75000/0.75000 \\ 
&4000 & 0.00500/0.50000 & 0.50000/0.50000 \\ 
&8000 & 0.25000/0.25000 & 0.25000/0.25000 \\ 
&16000 & 0.10000/0.25000 & 0.25000/0.25000 \\ 
&32000 & 0.07500/0.10000 & 0.25000/0.25000 \\  
\end{tabular}
\caption{}
\end{subtable}
}
\caption{Minimum and maximum tuning parameter values (min/max) picked by (a) the CMI and L1LR method in Section \ref{sec:SYNTH_SIM}, and (b) the CMI method in Section \ref{sec:BM_SIM}. \label{tab:TH_REG}}
\end{centering}
\end{sidewaystable}

\clearpage
\bibliography{MPLbib}

\end{document}